# Symmetry, Saddle Points, and Global Optimization Landscape of Nonconvex Matrix Factorization

X. Li, J. Lu, R. Arora, J. Haupt, H. Liu, Z. Wang, and T. Zhao*

**Abstract**

We propose a general theory for studying the landscape of nonconvex optimization with underlying symmetric structures for a class of machine learning problems (e.g., low-rank matrix factorization, phase retrieval, and deep linear neural networks). In specific, we characterize the locations of stationary points and the null space of Hessian matrices of the objective function via the lens of invariant groups. As a major motivating example, we apply the proposed general theory to characterize the global landscape of the nonconvex optimization in low-rank matrix factorization problem. In particular, we illustrate how the rotational symmetry group gives rise to infinitely many nonisolated strict saddle points and equivalent global minima of the objective function. By explicitly identifying all stationary points, we divide the entire parameter space into three regions: ($\mathcal{R}_1$) the region containing the neighborhoods of all strict saddle points, where the objective has negative curvatures; ($\mathcal{R}_2$) the region containing neighborhoods of all global minima, where the objective enjoys strong convexity along certain directions; and ($\mathcal{R}_3$) the complement of the above regions, where the gradient has sufficiently large magnitudes. We further extend our result to the matrix sensing problem. Such global landscape implies strong global convergence guarantees for popular iterative algorithms with arbitrary initial solutions.

## 1 Introduction

We consider a low-rank matrix estimation problem. Specifically, we want to estimate $M^* \in \mathbb{R}^{n \times m}$ with $\text{rank}(M^*) = r \ll \min\{n, m\}$ by solving the following rank-constrained problem

$$\min_M f(M) \quad \text{subject to } \text{rank}(M) \leq r, \tag{1}$$

*Xingguo Li and Jarvis Haupt are affiliated with Department of Electrical and Computer Engineering at University of Minnesota, Minneapolis, MN, 55455, USA; Raman Arora is affiliated with Department of Computer Science at Johns Hopkins University Baltimore, MD, 21210, USA; Zhaoran Wang, Junwei Lu and Han Liu are affiliated with Department of Operations Research and Financial Engineering at Princeton University, Princeton, NJ 08544, USA; Tuo Zhao is affiliated with School of Industrial and Systems Engineering at Georgia Institute of Technology, Atlanta, GA 30332; Emails: `lixx1661@umn.edu, tourzhao@gatech.edu`; Tuo Zhao is the corresponding author. This research is supported by DARPA Young Faculty Award N66001-14-1-4047; NSF DMS-1454377-CAREER; NSF IIS-1546482-BIGDATA; NIH R01MH102339; NSF IIS-1408910; NSF IIS-1332109; NIH R01GM083084.

where $f : \mathbb{R}^{n \times m} \to \mathbb{R}$ is usually a convex and smooth loss function. Since solving (1) has been known to be NP-hard in general, significant efforts have been also devoted to studying a convex relaxation of (1) as follows,

$$\min_{M} f(M) \quad \text{subject to} \quad \|M\|_* \leq \tau, \tag{2}$$

where $\tau$ is a tuning parameter and $\|M\|_*$ is the sum of all singular values of $M$, also known as the nuclear norm Candès and Recht (2009); Recht et al. (2010); Cai et al. (2010); Becker et al. (2011).

Although there have been a number of algorithms proposed for solving either (1) or (2) in existing literature Jain et al. (2010); Lee and Bresler (2010); Shalev-Shwartz et al. (2011), all these algorithms are iterative, and each iteration needs to calculate a computationally expensive Singular Value Decomposition (SVD), or an equivalent operation for finding the dominant singular values/vectors. This is very prohibitive for large-scale problems. In practice, most of popular heuristic algorithms resort to factorizing $M$ to a product of smaller matrices, i.e, $M^* = UV^\top$, where $U \in \mathbb{R}^{n \times r}$ and $V \in \mathbb{R}^{m \times r}$, also known as the factorized form. Then instead of solving (1) or (2), we solve the following nonconvex problem

$$\min_{X \in \mathbb{R}^{n \times r}, Y \in \mathbb{R}^{m \times r}} f(XY^\top), \tag{3}$$

where scalable algorithms can iteratively update $X$ and $Y$ very efficiently. The reparametrization of the low rank matrix in (3) is closely related to the Burer-Monteiro factorization for semidefinite programing in existing literature. See more details in Burer and Monteiro (2003, 2005).

Tremendous progress has been made to provide theoretical justifications of the popular nonconvex factorization heuristic algorithms for general classes of functions Bhojanapalli et al. (2016a); Ge et al. (2015); Chen and Wainwright (2015); Anandkumar and Ge (2016); Park et al. (2016a). A wide family of problems can be cast as (3). Popular examples include matrix sensing Bhojanapalli et al. (2016a); Zhao et al. (2015); Tu et al. (2015); Chen and Wainwright (2015); Bhojanapalli et al. (2016b); Park et al. (2016b), matrix completion Keshavan et al. (2009); Hardt (2014); Sun and Luo (2015); Zheng and Lafferty (2016); Ge et al. (2016); Jin et al. (2016), sparse principle component analysis (PCA) Cai et al. (2013); Birnbaum et al. (2013); Vu et al. (2013), and factorization machine Lin and Ye (2016); Blondel et al. (2016). Recent efforts are also made when the observation is a superposition of low-rank and sparse matrices Yi et al. (2016); Gu et al. (2016). Moreover, extensions to low-rank tensor estimation and its related problems, such as independent component analysis (ICA) and topic modeling, are also studied Ge et al. (2016); Arias-Castro and Verzelen (2014); Arora et al. (2012); Zou et al. (2013).

The factorized form $M = XY^\top$ makes (3) very challenging to solve. First, it yields infinitely many nonisolated saddle points because of the existence of invariant rotation group. For example, if some $(X, Y)$ pair is a saddle point, then for any orthogonal matrix $\Phi \in \mathbb{R}^{r \times r}$, i.e., $\Phi\Phi^\top = I$, $(X\Phi, Y\Phi)$ is also a saddle point since $XY^\top = X\Phi(Y\Phi)^\top$. For the same reason, there exist infinitely many local/global minima as well for $r > 1$. Second, although $f(M)$ is convex on $M$, $f(XY^\top)$ is not jointly convex in $X$ and $Y$ (even around a small neighborhood of a global optimum). To address these challenges, various techniques are developed recently. Extensive contemporary works focus on the local convergence rate analysis based on local geometric properties of the optimization problem using generalization of convexity/smoothness of $f$, such as local regularity condition (Tu et al., 2015; Zheng and Lafferty, 2016, 2015; Candes et al., 2015) and local descent condition (Chen and Wainwright, 2015; Yi et al., 2016). However, careful initialization is required in this type of approaches. Another line of works on solving the factorized problem (3) focus on the optimality conditions that guarantees global convergence using random initialization Bhojanapalli et al. (2016b); Ge et al. (2016). However, since only partial results on the landscape of optimization are discussed, e.g., only stationary points (i.e., saddle points and local minima) are characterized without discussing their neighborhood or the rest region of the parameter space, no explicit global convergence rate can be guaranteed.

In addition to the approaches discussed above, another more clear yet more challenging scheme is to characterize the global landscape of the nonconvex optimization problem, based on which the global convergence analysis becomes possible. Without further distinction, we use "landscape" to denote the the geometry of the objective function in the optimization problem, i.e., *the characterization of all stationary points* and *the explicit geometry of the objective function on the entire parameter domain* (e.g., the characterization of regions $\mathcal{R}_1$, $\mathcal{R}_2$, and $\mathcal{R}_3$ defined below). Nevertheless, there are few works that discuss the global landscape of the nonconvex optimization (3) in such an explicit manner. One of the earliest works that study the global landscape of nonconvex optimization in this sense is on the phase retrieval problem (Sun et al., 2016), which can be viewed as a special case of (3). Such global landscape on optimization can further help provide global convergence rate analysis using popular iterative algorithms without careful initialization (Ge et al., 2015; Sun et al., 2016; Lee et al., 2016; Panageas and Piliouras, 2016). However, existing works have not discussed the intrinsic reasons of difficulties that present in the nonconvex matrix factorization problems, e.g., the generation of saddle points.

To shed light on the nonconvex matrix factorization problems (3), our study in this paper consists of two major parts to answer two questions of our interest: (I) Why are there saddle points and how to identify them effectively? (II) How do the saddle points impact the geometry of the optimization problem? To answer the first question, we study a generic theory for characterizing the landscape of a general class of functions with underlying symmetric structures. Based on a new symmetry principle, we identify stationary points for those functions with invariant groups, which characterizes the underlying principle of generating saddle points in nonconvex matrix factorization problems. Moreover, we characterize the null space of the Hessian matrices of the stationary points via the tangent space. We further provide concrete examples to demonstrate our proposed theory. To the best of our knowledge, this is the first effort to provide a generic framework for characterizing geometric properties of a large class of functions with symmetric structure.

To answer the second question, we establish a comprehensive analysis for global landscape of the low-rank matrix factorization problem based on our proposed generic theory. Specifically, we consider a symmetric positive semidefinite (PSD) matrix $M^* = UU^\top \succeq 0$, and solve the following problem

$$\min_{X \in \mathbb{R}^{n \times r}} \mathcal{F}(X), \text{ where } \mathcal{F}(X) = \frac{1}{4} \|M^* - XX^\top\|_F^2. \tag{4}$$

Here we only consider the PSD matrix for simplicity, and the extension to the general rectangular case is straightforward (see more details in Section 2). Though (4) has been viewed as an important foundation of many popular matrix factorization problems such as matrix sensing and matrix completion, the global landscape of $\mathcal{F}(X)$ in (4) is not very clear yet. Based on our generic theory, we explicitly identify all saddle points and global minima of $\mathcal{F}(X)$. Further, we show that the entire parameter space can be described as one the three regions as follows.

($\mathcal{R}_1$) The region that contains neighborhoods of all saddle points, where any associated Hessian matrix of the objective has negative eigenvalues. This so-called strict saddle property guarantees that many commonly used iterative algorithms cannot not be trapped in those saddle points.

($\mathcal{R}_2$) The region that contains neighborhoods of all global minima, where the objective is only strongly convex along certain trajectories, otherwise is nonconvex, unless $r = 1$. We specify these directions explicitly, along which $\mathcal{F}(X)$ is strongly convex.

($\mathcal{R}_3$) The complement of regions $\mathcal{R}_1$ and $\mathcal{R}_2$ in $\mathbb{R}^{n \times r}$, where the gradient has a sufficiently large norm. Together with $\mathcal{R}_1$ and $\mathcal{R}_2$, a convergence of (4) to a global minimum is guaranteed for many commonly used iterative algorithms without special initializations.

Moreover, we further connect our analysis on (4) to the matrix sensing problem, which can be considered as a perturbed version of (4). Using a suboptimal sample complexity, we establish analogous global geometric properties to (4) for the matrix sensing problem. These strong geometric properties imply the convergence to a global minimum of the matrix factorization problem in polynomial time without careful initialization for several popular iterative algorithms, such as the gradient descent algorithm, the noisy stochastic gradient descent algorithm, and the trust-region Newton's algorithm.

After the initial release of our paper, several concurrent and follow-up works have appeared. In specific, Zhu et al. (2017) extend our analysis to the general rectangular matrices using the lifting formulation and achieve analogous results to ours. Another related work is Ge et al. (2017), which provide a unified geometric analysis based on the strict saddle property for several popular nonconvex problems, including matrix sensing, matrix completion, and robust PCA. By partially applying the result in Ge et al. (2017), we further demonstrate a sharper result for matrix sensing in terms of the sample complexity, with some sacrifice in the properties of the optimization landscape as a tradeoff. Further discussions will be provided in Section 3.3 and 5.1.

The rest of the paper is organized as follows. In Section 2, we provide a generic theory of identifying stationary points and the null space of their Hessian matrices, along with several concrete examples. In Section 3, a global geometric analysis is established for the low-rank matrix factorization problem. In Section 4, we extend the analysis to the matrix sensing problem, followed by a further discussion in Section 5. All proofs are deferred to Appendix.

**Notation**. Given an integer $n \geq 1$, we denote $[n] = \{1, \ldots n\}$. Let $\mathcal{O}_r = \{\Psi \in \mathbb{R}^{r \times r} \mid \Psi \Psi^\top = \Psi^\top \Psi = I_r\}$ be the set of all orthogonal matrices in $\mathbb{R}^{r \times r}$. Given a matrix $A \in \mathbb{R}^{n \times m}$ and a subspace $\mathcal{L} \in \mathbb{R}^n$, let $\mathcal{P}_\mathcal{L}(A)$ be the orthogonal projection operation of $A$ onto $\mathcal{L}$, and $\mathcal{L}^\perp$ be the complement of $\mathcal{L}$ in $\mathbb{R}^n$. Denote $\mathcal{L}_A$ as the column space of $A$. We use $A_{(*,k)}$ and $A_{(j,*)}$ to denote the $k$-th column and the $j$-th row respectively, $A_{(j,k)}$ to denote the $(j,k)$-th entry, and $A_\mathcal{S}$ to denote a column-wise sub matrix of $A$ indexed by a set $\mathcal{S} \subseteq [m]$. Let $\sigma_i(A)$ be the $i$-th largest singular value, $\|A\|_2$ be the spectral norm (largest singular value), and $\|A\|_F$ be the Frobenius norm. Given two matrices $A, B \in \mathbb{R}^{n \times m}$, denote $\langle A, B \rangle = \text{Tr}(A^\top B) = \sum_{i,j} A_{(i,j)} B_{(i,j)}$. When $A \in \mathbb{R}^{n \times n}$ is a square matrix, we denote $\lambda_{\max}(A)$ and $\lambda_{\min}(A)$ as the largest and smallest eigenvalues respectively. Given a vector $a \in \mathbb{R}^n$, let $a_{(i)}$ be the $i$-th entry. We use a subscript $A_i$ ($a_i$) to denote the $i$-th matrix (vector) in a sequence of matrices (vectors). Denote $\mathbb{E}(X)$ as the expectation of a random variable $X$ and $\mathbb{P}(\mathcal{X})$ as the probability of an event $\mathcal{X}$. We use $\otimes$ as the kronecker product, and preserve $C_1, C_2, \ldots$ and $c_1, c_2, \ldots$ for positive real constants.

## 2  A Generic Theory for Stationary Points

Given a function $f$, our goal is to find the stationary point. Rigorous mathematical definitions are provided as follows.

**Definition 1.** Given a smooth function $f : \mathbb{R}^n \to \mathbb{R}$, a point $x \in \mathbb{R}^n$ is called:

(i) a ***stationary point***, if $\nabla f(x) = 0$;

(ii) a ***local minimum (or maximum)***, if $x$ is a stationary point and there exists a neighborhood $\mathcal{B} \subseteq \mathbb{R}^n$ of $x$ such that $f(x) \leq f(y)$ (or $f(x) \geq f(y)$) for any $y \in \mathcal{B}$;

(iii) a ***global minimum (or maximum)***, if $x$ is a stationary point and $f(x) \leq f(y)$ (or $f(x) \geq f(y)$) for any $y \in \mathbb{R}^n$;

(iv) a ***strict saddle point***, if $x$ is a stationary point and for any neighborhood $\mathcal{B} \subseteq \mathbb{R}^n$ of $x$, there exist $y, z \in \mathcal{B}$ such that $f(z) < f(x) < f(y)$ and $\lambda_{\min}(\nabla^2 f(x)) < 0$.

A visualization of different types of stationary points are provided in Figure 1. In general, finding the stationary point requires solving a large system $\nabla f(x) = 0$, which can be computationally challenging. However, when $f$ has special structures, we can develop new principles to find the set of stationary points conveniently.

In this paper, we consider a class of functions with invariant groups, for which we provide a generic theory to determine the stationary point using the symmetry principle. This covers the low-rank matrix factorization problem as a special example. Moreover, we can characterize the null space of the Hessian matrix at the stationary point by leveraging the tangent space. This will further help us to determine the saddle point and local/global minimum (see more details in Section 3).

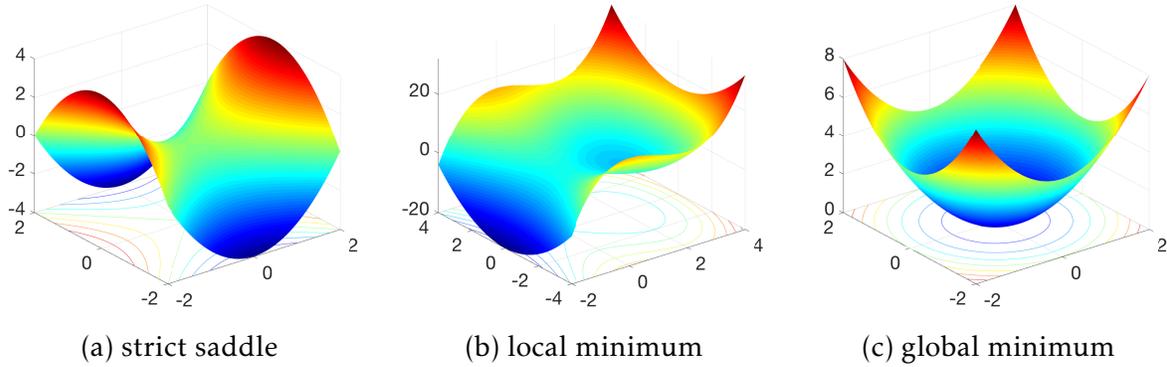

(a) strict saddle  (b) local minimum  (c) global minimum

Figure 1: Examples of a strict saddle point, a local minimum, and a global minimum.

## 2.1 Determine Stationary Points

For self-containedness, we start with a few definitions in group theory Dummit and Foote (2004) as follows.

**Definition 2.** A ***group*** $\mathcal{G}$ is a set of elements together with a binary operation $\{\cdot\}$ that satisfies the following four properties:

- Closure: for all $a_1, a_2 \in \mathcal{G}$, we have $a_1 \cdot a_2 \in \mathcal{G}$;
- Associativity: for all $a_1, a_2, a_3 \in \mathcal{G}$, we have $(a_1 \cdot a_2) \cdot a_3 = a_1 \cdot (a_2 \cdot a_3)$;
- Identity: there exists an identity element $e \in \mathcal{G}$ such that $e \cdot a = a$ and $a \cdot e = a$ for all $a \in \mathcal{G}$;
- Inverse: for any $a \in \mathcal{G}$, there exists an inverse element $a^{-1} \in \mathcal{G}$ such that $a \cdot a^{-1} = e$ and $a^{-1} \cdot a = e$.

**Definition 3.** A ***commutative group*** is a group that also satisfies

- Commutativity: for all $a_1, a_2 \in \mathcal{G}$, we have $a_1 \cdot a_2 = a_2 \cdot a_1$.

**Definition 4.** A ***field*** is a set with two binary operations $\{+, \cdot\}$, addition (denoted $\{+\}$ and multiplication $\{+, \cdot\}$, both of which satisfy associativity, identity ($\{+\}$ is associated with identity 0 and $\{\cdot\}$ is associated with identity 1), inverse, commutativity, and

- Distributivity: for all $a_1, a_2, a_3 \in \mathcal{G}$, we have $a_1 \cdot (a_2 + a_3) = (a_1 \cdot a_2) + (a_1 \cdot a_3)$.

**Definition 5.** A subset $\mathcal{H}$ of a group $\mathcal{G}$ is a ***subgroup*** if $\mathcal{H}$ is itself a group under the operation induced by $\mathcal{G}$.

**Definition 6.** The set of all invertible $n \times n$ real matrices with determinant 1, together with the operations of ordinary matrix multiplication and matrix inversion, is a ***special linear group*** of degree $n$ over a field, denoted as $\mathrm{SL}_n(\mathbb{R})$.

**Definition 7.** Given a function $f : \mathbb{R}^m \to \mathbb{R}$, a subgroup $\mathcal{G}$ of a special linear group $\mathrm{SL}_m(\mathbb{R})$ is an ***invariant group*** if $\mathcal{G}$ satisfies $f(x) = f(g(x))$ for all $x \in \mathbb{R}^m$ and $g \in \mathcal{G}$.

**Remark 1.** We define the invariant group in terms of the special linear group rather than the general linear group because we want to preserve the volume for linear transformations.

**Definition 8.** A point $x_\mathcal{G}$ is a ***fixed point*** of a group $\mathcal{G}$ if $g(x_\mathcal{G}) = x_\mathcal{G}$ for all $g \in \mathcal{G}$.

**Definition 9.** Given a linear space $\mathcal{X}$, let $\mathcal{Y}$ and $\mathcal{Z}$ be subspaces of $\mathcal{X}$. Then $\mathcal{X}$ is the ***direct sum*** of $\mathcal{Y}$ and $\mathcal{Z}$, denoted as $\mathcal{X} = \mathcal{Y} \oplus \mathcal{Z}$, if we have $\mathcal{X} = \{y + z \mid y \in \mathcal{Y}, z \in \mathcal{Z}\}$ and $\mathcal{Y} \cap \mathcal{Z} = \{0\}$.

Note that the direct sum we used throughout this paper is the *internal direct sum* since $\mathcal{Y}$ and $\mathcal{Z}$ are subspaces rather than spaces. We then present a generic theory of determining stationary points as follows. The proof is provided in Appendix A.1.

**Theorem 1** (Stationary Fixed Point)**.** Suppose $f$ has an invariant group $\mathcal{G}$ and $\mathcal{G}$ has a fixed point $x_\mathcal{G}$. If we have

$$\mathcal{G}(\mathbb{R}^m) \triangleq \mathrm{Span}\{g(x) - x \mid g \in \mathcal{G}, x \in \mathbb{R}^m\} = \mathbb{R}^m,$$

then $x_\mathcal{G}$ is a stationary point of $f$.

By Theorem 1, we can find a stationary point of functions with invariant groups given a fixed point. Refined result can be obtained for subspaces when we consider a decomposition $\mathbb{R}^m = \mathcal{Y} \oplus \mathcal{Z}$, where $\mathcal{Y}$ and $\mathcal{Z}$ are orthogonal subspaces of $\mathbb{R}^m$. This naturally induces a subgroup of $\mathcal{G}$ as

$$\mathcal{G}_\mathcal{Y} = \{g_\mathcal{Y} \mid g_\mathcal{Y}(y) = g(y \oplus 0), g \in \mathcal{G}, y \in \mathcal{Y}, 0 \in \mathcal{Z}\}.$$

Obviously, $\mathcal{G}_\mathcal{Y}$ is a subgroup of a special linear group on $\mathcal{Y}$. Moreover, $y_{\mathcal{G}_\mathcal{Y}} = \mathcal{P}_\mathcal{Y}(x_\mathcal{G}) \in \mathcal{Y}$ is a fixed point of $\mathcal{G}_\mathcal{Y}$, where $\mathcal{P}_\mathcal{Y}$ is a projection operation onto $\mathcal{Y}$. We then have the following corollary immediately from Theorem 1.

**Corollary 1.** If $y_{\mathcal{G}_\mathcal{Y}}$ is a fixed point of $\mathcal{G}_\mathcal{Y}$ and

$$z^*(y_{\mathcal{G}_\mathcal{Y}}) \in \arg\mathrm{zero}_z \nabla_z f(y_{\mathcal{G}_\mathcal{Y}} \oplus z),$$

where $\arg\mathrm{zero}_z \nabla_z f(y_{\mathcal{G}_\mathcal{Y}} \oplus z)$ is the set of zero solutions of $\nabla_z f(y \oplus z)$ by fixing $y = y_{\mathcal{G}_\mathcal{Y}}$, then $g(y_{\mathcal{G}_\mathcal{Y}} \oplus z^*)$ is a stationary point for all $g \in \mathcal{G}$.

Given a fixed point in a subspace, we have from Corollary 1 that the direct sum of the fixed point and any zero solution of the partial derivative of the function with respect to the orthogonal subspace is also a stationary point. This allows us to recursively use Theorem 1 and Corollary 1 to find a set of stationary points. We call such a procedure the *symmetry principle* of stationary point. Here, we demonstrate some popular examples with symmetric structures.

**Example 1** (Low-rank Matrix Factorization). Recall that given a PSD matrix $M^* = UU^\top$ for some $U \in \mathbb{R}^{n \times r}$, the objective function with respect to variable $X \in \mathbb{R}^{n \times r}$ admits

$$f(X) = \frac{1}{4}\|XX^\top - M^*\|_F^2. \tag{5}$$

Given $g = \Psi_r \in \mathcal{O}_r$, let $g(X) = X\Psi_r$, then we have $f(X) = f(g(X))$. It is easy to see that the rotation group $\mathcal{G} = \mathcal{O}_r$ is an invariant group of $f$ and $X_\mathcal{G} = 0$ is a fixed point. Theorem 1 implies that 0 is a stationary point.

The gradient of $f(X)$ is

$$\nabla f(X) = (XX^\top - M^*)X.$$

We consider the subspace $\mathcal{Y} \subseteq \mathcal{L}_U$ of the column space of $U$ and $X_{\mathcal{G}_\mathcal{Y}} = 0_\mathcal{Y}$. Applying Corollary 1 to $\mathcal{Y} = \{0\}$ and $\mathcal{Z} = \mathcal{L}_U$, we have $U\Psi_r$ is a stationary point, where $\Psi_r \in \mathcal{O}_r$. Analogously, applying Corollary 1 again to $\mathcal{Y} = \mathcal{L}_{U_{r-s}} \subseteq \mathcal{L}_U$ and $\mathcal{Z} = \mathcal{L}_{U_s} \subseteq \mathcal{L}_U$, we have $U_s\Psi_r$ is a stationary point of $f(X)$, where $\Psi_r \in \mathcal{O}_r$, $U_s = \Phi\Sigma S\Theta^\top$ and $U_{r-s} = \Phi\Sigma(I - S)\Theta^\top$ given the SVD of $U = \Phi\Sigma\Theta^\top$, and $S$ is a diagonal matrix with arbitrary $s$ entries being 1 and the rest being 0 for all $s \in [r]$. This will be discussed in further details in Section 3. Note that the degree of freedom of $\Psi_r$ in $U_s\Psi_r$ is in fact $s(s-1)/2$ instead of $r(r-1)/2$, since $U_s$ is of rank $s$.

The result can be easily extended to general low-rank rectangular matrices. For $X, U \in \mathbb{R}^{n \times r}$ and $Y, V \in \mathbb{R}^{m \times r}$, we consider the function

$$f(X, Y) = \frac{1}{2}\|XY^\top - M^*\|_F^2. \tag{6}$$

Using the similar analysis for the symmetric case above, we have $(X, Y) = (0, 0)$ and $(X, Y) = (U\Psi_r, V\Psi_r)$ are both stationary points. Moreover, given the SVD of $UV^\top = \Phi\Sigma\Theta^\top$, we have $(X, Y) = (\Phi\Sigma_1 S\Psi_r, \Theta\Sigma_2 S\Psi_r)$ is a stationary point, where $\Sigma_1\Sigma_2 = \Sigma$, and $S$ is a diagonal matrix with arbitrary $s$ entries being 1 and the rest being 0, for all $s \in [r]$.

**Example 2** (Phase Retrieval). Given i.i.d. complex Gaussian vectors $\{a_i\}_{i=1}^m$ in $\mathbb{C}^n$ and measurements $y_i = |a_i^H u|$ of complex vector $u \in \mathbb{C}^n$ for $i = 1, \ldots, m$, where $x^H$ is the Hermitian transpose, a natural square error formulation of the objective of phase retrieval with respect to variable $x \in \mathbb{C}^n$ Candes et al. (2015); Sun et al. (2016) is

$$h(x) = \frac{1}{2m}\sum_{i=1}^m \left(y_i^2 - |a_i^H x|^2\right)^2.$$

For simplicity, we consider the expected objective of $h$ as

$$f(x) = \mathbb{E}(h(x)) = \|x\|_2^4 + \|u\|_2^4 - \|x\|_2^2\|u\|_2^2 - |x^H u|^2,$$

It is easy to see that $f$ has an invariant group $\mathcal{G} = \left\{e^{i\theta} \mid \theta \in [0, 2\pi)\right\}$ and $x_\mathcal{G} = 0$ is a fixed point. Then Theorem 1 implies that 0 is a stationary point.

The gradient of $f(x)$ is

$$\nabla f(x) = \begin{bmatrix} (2\|x\|_2^2 I - \|u\|_2^2 I - uu^H)x \\ (2\|x\|_2^2 I - \|u\|_2^2 I - uu^H)\bar{x} \end{bmatrix},$$

where $\bar{x}$ is the complex conjugate. Consider a coordinate-wise subspace $\mathcal{Y} \subseteq \mathbb{C}^n$ of degree $k \leq n$, where for any $\widetilde{y} \in \mathcal{Y}$, $\widetilde{y}$ shares identical entire with $x$ in certain $k$ coordinates and has zero entries otherwise. Applying Corollary 1 to $\mathcal{Y} = \{0\}$, i.e., $k = 0$, we have that $ue^{i\theta}$ is a stationary point for any $\theta \in [0, 2\pi)$. For $\mathcal{Y} \neq \{0\}$, i.e., $k > 0$, we have $z^*(0_{\mathcal{Y}}) \in \mathcal{D} = \{x \in \mathcal{Z} \mid x^H u = 0, \ x_{\mathcal{Y}} = 0, \ \|x\|_2 = \|u\|_2/\sqrt{2}\}$. Applying Corollary 1 again, we have $xe^{i\theta}$ is a stationary point for any $x \in \mathcal{D}$ and $\theta \in [0, 2\pi)$.

**Example 3** (Deep Linear Neural Networks). Given data $W \in \mathbb{R}^{n_0 \times m}$ and $Y \in \mathbb{R}^{n_L \times m}$, we consider a square error objective of a feedforward deep linear neural network of $L$ layers Goodfellow et al. (2016),

$$f(X_1, \ldots, X_L) = \frac{1}{2} \|X_L X_{L-1} \cdots X_1 W - Y\|_F^2,$$

where $X_l \in \mathbb{R}^{n_l \times n_{l-1}}$ is the weight matrix in the $l$-th layer for all $l \in [L]$. We can see that for any $l \in [L-1]$, $f$ has orthogonal groups $\mathcal{G}_l = \mathcal{O}_{n_l}$ as the invariant groups and $X_{\mathcal{G}_l} = 0$ is a fixed point. Theorem 1 implies that 0 is a stationary point.

The blockwise structure naturally leads to a derivation of further stationary points by fixing all but one block. Specifically, given some $l \in [L-1]$, we fix all the other blocks $[L-1]\setminus\{l\}$, then the gradient of $f(X_1, \ldots, X_L)$ with respect to $X_l$ is

$$\nabla_{X_l} f(X_1, \ldots, X_L) = A^\top (A X_l B - Y) B^\top,$$

where $A = X_L \cdots X_{l+1}$ and $B = X_{l-1} \cdots X_1 W$. We consider a pair $(X_{l+1}, X_l)$. Solving $\nabla_{X_l} f(X_1, \ldots, X_L) = 0$, $X_l$ is a fixed point if $X_l$ satisfies

$$X_l = (A^\top A)^- A^\top Y B^\top (BB^\top)^- + (I - (A^\top A)^- A^\top A)Q,$$

where $D^-$ is a generalized inverse of the matrix $D$ and $Q \in \mathbb{R}^{n_l \times n_{l-1}}$ is an arbitrary matrix. We consider a subspace $\mathcal{Y} \subset \mathcal{L}_{I-(A^\top A)^- A^\top A}$, then Corollary 1 implies that $(X_{l+1}\Psi_{n_l}, \Psi_{n_l}^\top X_l)$ is a pair of stationary point, where $X_l \in \mathcal{Y}$ and $\Psi_{n_l} \in \mathcal{O}_{n_l}$. Moreover, for $\mathcal{Y} = \mathcal{L}_{I-(A^\top A)^- A^\top A}$, we have $z^*(0_{\mathcal{Y}}) = (A^\top A)^- A^\top Y B^\top (BB^\top)^-$ and Corollary 1 implies that $\left(X_{l+1}\Psi_{n_l}, \Psi_{n_l}^\top (A^\top A)^- A^\top Y B^\top (BB^\top)^-\right)$ is also a pair of stationary point, where $\Psi_{n_l} \in \mathcal{O}_{n_l}$.

## 2.2 Null Space of Hessian Matrix at Stationary Points

We now discuss the null space of the Hessian matrix at a stationary point, which can be used to further distinguish between saddle point and local/global minimum. Our intuition is that the null space of the Hessian matrix should contain the vectors tangent to the invariant group $\mathcal{G}$. We start with a few definitions in manifold Robbin and Salamon (2011) as follows.

**Definition 10** (Manifold). Given positive integers $m$ and $k$, we call a subset $\mathcal{M} \subset \mathbb{R}^m$ as a **smooth $k$-dimensional manifold** (or a **smooth $k$-submanifold**) if every point $x \in \mathcal{M}$ has an open neighborhood $\mathcal{X} \subset$ such that $\mathcal{X} \cap \mathcal{M}$ is diffeomorphic to an open subset $\mathcal{B} \subset \mathbb{R}^k$, i.e., there exists a function $f : \mathcal{X} \cap \mathcal{M} \to \mathcal{B}$ such that $f$ is bijective, and $f$ and $f^{-1}$ are smooth.

**Definition 11** (Tangent Space). Let $\mathcal{M} \subset \mathbb{R}^m$ be a smooth $k$-dimensional manifold. Given $x \in \mathcal{M}$, we call $v \in \mathbb{R}^m$ as a **tangent vector** of $\mathcal{M}$ at $x$ if there exists a smooth curve $\gamma : \mathbb{R} \to \mathcal{M}$ with $\gamma(0) = x$ and $v = \gamma'(0)$. The set of tangent vectors of $\mathcal{M}$ at $x$ is called the **tangent space** of $\mathcal{M}$ at $x$, denoted as

$$T_x\mathcal{M} = \{\gamma'(0) \mid \gamma : \mathbb{R} \to \mathcal{M} \text{ is smooth }, \gamma(0) = v\}.$$

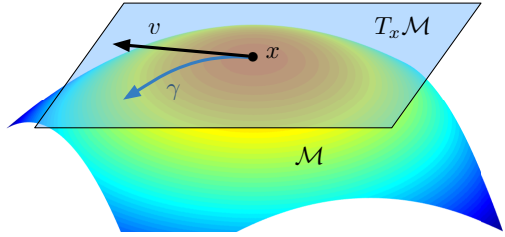

Figure 2: A graphical illustration of a manifold $\mathcal{M}$ and a tangent space $T_x\mathcal{M}$ at some point $x$ on the manifold. $v$ is a tangent vector at $x$ and $\gamma$ is the corresponding smooth curve.

A visualization of the manifold and the tangent space is provided in Figure 2. The following theorem shows that the null space of the Hessian matrix at a stationary point $x$ contains the tangent space of the set $\mathcal{G}(x) = \{g(x) \mid g \in \mathcal{G}\}$. The proof is provided in Appendix A.2.

**Theorem 2.** *If $f$ has an invariant group $\mathcal{G}$ and $H_x$ is the Hessian matrix at a stationary point $x$, then we have*

$$T_x\mathcal{G}(x) \subseteq \mathrm{Null}(H_x).$$

In the following, we demonstrate examples discussed in Section 2.1 to instantiate Theorem 2.

**Example 4** (Low-rank Matrix Factorization). Remind that for low-rank matrix factorization in Example 1, $f$ has an invariant group $\mathcal{G} = \mathcal{O}_r$, which is also a smooth submanifold in $\mathbb{R}^{r \times r}$ of dimension $r(r-1)/2$. Given any $X \in \mathbb{R}^{n \times r}$, let $\gamma : \mathbb{R} \to \mathcal{O}_r(X)$ be a smooth curve, i.e., for every $t \in \mathbb{R}$ there exists $\Psi_r \in \mathcal{O}_r$ such that $\gamma(t) = g_t(X) = X\Psi_r$ and $\gamma(0) = g_0(X) = X$. By definition, for any $t \in \mathbb{R}$, we have

$$\gamma(t)\gamma(t)^\top = XX^\top.$$

Differentiating both sides, we have $\gamma'(t)\gamma(t)^\top + \gamma(t)\gamma'(t)^\top = 0$. Plugging in $t = 0$, we have

$$\gamma'(0)X^\top + X\gamma'(0)^\top = 0.$$

Then we can see that

$$T_X \mathcal{O}_r(X) = \{XE \mid E \in \mathbb{R}^{r \times r}, E = -E^\top\}.$$

By Example 1, we have that $U_s \Psi_r$ is a stationary point for $\mathcal{Y} = \mathcal{L}_{U_{r-s}} \subseteq \mathcal{L}_U$. Theorem 2 implies that for any skew symmetric matrix $E \in \mathbb{R}^{r \times r}$, we have $U_s \Psi_r E$ belongs to the null space of the Hessian matrix at $U_s \Psi_r$. Similar to $\Psi_r$, the dimension of $T_X \mathcal{O}_r(X)$ at $X = U_s \Psi_r E$ depends on $s$ since $U_s$ is of rank $s$. Specifically, the dimension of the tangent space is at least $s(s-1)/2 + (n-(r-s))(r-s)$, where $s(s-1)/2$ is the degree of freedom of the set of $E$ and $(n-(r-s))(r-s)$ is degree of freedom of $U_s \Psi_r$.

**Example 5** (Phase Retrieval). For phase retrieval in Example 2, $f$ has an invariant group $\mathcal{G} = \{e^{i\theta} \mid \theta \in [0, 2\pi)\}$. Given any $x \in \mathbb{C}^n$, let $\gamma : \mathbb{R} \to \mathcal{G}(x)$ be a smooth curve, i.e., for every $t \in \mathbb{R}$ there exists $\theta \in [0, 2\pi)$ such that $\gamma(t) = x e^{i\theta}$ and $\gamma(0) = x$. Then for any $t \in \mathbb{R}$, we have

$$\|\gamma(t)\|_2^2 = \|x\|_2^2.$$

Differentiating both sides, we have $\gamma'(t)^H \gamma(t) + \gamma(t)^H \gamma'(t) = 0$. Plugging in $t = 0$, we have

$$\gamma'(0)^H x = -x^H \gamma'(0).$$

Then we can see that

$$T_x \mathcal{G}(x) = ix.$$

By Example 2, we have $ue^{i\theta}$ is a stationary point for all $\theta \in [0, 2\pi)$. Theorem 2 implies that $iue^{i\theta}$ belongs to the null space of Hessian matrix at $ue^{i\theta}$.

**Example 6** (Deep Linear Neural Networks). For the deep linear neural networks in Example 3, $f$ has an invariant group $\mathcal{G}_l = \mathcal{O}_{n_l}$ for any $l \in [L-1]$. Using the same analysis in Example 4, we have that for any skew symmetric matrix $E \in \mathbb{R}^{r \times r}$, the pair $\left(X_{l+1} \Psi_{n_l} E, E^\top \Psi_{n_l}^\top (A^\top A)^- A^\top Y B^\top (BB^\top)^-\right)$ belongs to the null space of Hessian matrix for a stationary pair $\left(X_{l+1} \Psi_{n_l}, \Psi_{n_l}^\top (A^\top A)^- A^\top Y B^\top (BB^\top)^-\right)$.

## 3 A Geometric Analysis of Low-Rank Matrix Factorization

We apply our generic theories to study the global landscape of the low-rank matrix factorization problem. Our goal is to provide a comprehensive geometric perspective to fully characterize the low-rank matrix factorization problem (4). Finding all stationary points is the keystone, based on which we can further identify strict saddle points and global minima. This scheme has been adopted in geometry based convergence rate analyses to guarantee that iterative algorithms do not converge to the strict saddle point Ge et al. (2015); Sun et al. (2016); Lee et al. (2016); Panageas and Piliouras (2016). The landscape of the low-rank matrix factorization problem is also discussed briefly in Srebro and Jaakkola (2003), but no rigorous analysis is provided.

In particular, the zero of the gradient $\nabla \mathcal{F}(X)$ and the eigenspace of the Hessian matrix $\nabla^2 \mathcal{F}(X)$ are keys to our analysis. Given $\nabla \mathcal{F}(X)$ and $\nabla^2 \mathcal{F}(X)$, our analysis consists of the following major arguments:

(p1) identify all stationary points by finding the solutions of $\nabla \mathcal{F}(X) = 0$, which is further used to identify the strict saddle point and the global minimum,

(p2) identify the strict saddle point and their neighborhood such that $\nabla^2 \mathcal{F}(X)$ has a negative eigenvalue, i.e. $\lambda_{\min}(\nabla^2 \mathcal{F}(X)) < 0$,

(p3) identify the global minimum, their neighborhood, and the directions such that $\mathcal{F}(X)$ is strongly convex, i.e. $\lambda_{\min}(\nabla^2 \mathcal{F}(X)) > 0$, and

(p4) verify that the gradient has a sufficiently large norm outside the regions described in (p2) and (p3).

The analysis can be further extended to other problems, such as matrix sensing and matrix completion, which are considered as perturbed versions of (4). For simplicity, we first consider the PSD matrix $M^* = UU^\top$. Then we explain how to extend to a rectangular matrix, which is straightforward.

## 3.1 Warm-up: Rank 1 Case

We start with the basic case of $r = 1$ to obtain some insights. Specifically, suppose $M^* = uu^\top$, where $u \in \mathbb{R}^n$, then we consider

$$\min_{x \in \mathbb{R}^n} \mathcal{F}(x), \text{ where } \mathcal{F}(x) = \frac{1}{4}\|uu^\top - xx^\top\|_F^2. \tag{7}$$

The gradient and the Hessian matrix of $\mathcal{F}(x)$, respectively, are

$$\nabla \mathcal{F}(x) = (xx^\top - uu^\top)x \in \mathbb{R}^n \text{ and } \nabla^2 \mathcal{F}(x) = 2xx^\top + \|x\|_2^2 \cdot I_n - uu^\top \in \mathbb{R}^{n \times n}. \tag{8}$$

In the rank 1 case, the invariant group is $\mathcal{G} = \mathcal{O}_1 = \{1, -1\}$. We then provide the key arguments for the rank 1 setting in the following theorem. The proof is provided in Appendix B.

**Theorem 3.** Consider (7) and define the following regions:

$$\mathcal{R}_1 \triangleq \left\{ y \in \mathbb{R}^n \mid \|y\|_2 \leq \frac{1}{2}\|u\|_2 \right\},$$

$$\mathcal{R}_2 \triangleq \left\{ y \in \mathbb{R}^n \mid \min_{\psi \in \mathcal{O}_1} \|y - u\psi\|_2 \leq \frac{1}{8}\|u\|_2 \right\}, \text{ and}$$

$$\mathcal{R}_3 \triangleq \left\{ y \in \mathbb{R}^d \mid \|y\|_2 > \frac{1}{2}\|u\|_2, \min_{\psi \in \mathcal{O}_1} \|y - u\psi\|_2 > \frac{1}{8}\|u\|_2 \right\}.$$

Then the following properties hold.

(p1) $x = 0$, $u$ and $-u$ are the only stationary points of $\mathcal{F}(x)$.

(p2) $x = 0$ is a strict saddle point, where $\nabla^2 \mathcal{F}(0)$ is negative semi-definite with $\lambda_{\min}(\nabla^2 \mathcal{F}(0)) = -\|u\|_2^2$. Moreover, for any $x \in \mathcal{R}_1$, $\nabla^2 \mathcal{F}(x)$ contains a negative eigenvalue, i.e.

$$\lambda_{\min}(\nabla^2 \mathcal{F}(x)) \leq -\frac{1}{2}\|u\|_2^2.$$

(p3) For $x = \pm u$, $x$ is a global minimum, and $\nabla^2 \mathcal{F}(x)$ is positive definite with $\lambda_{\min}(\mathcal{F}(x)) = \|u\|_2^2$. Moreover, for any $x \in \mathcal{R}_2$, $\mathcal{F}(x)$ is locally strongly convex, i.e.

$$\lambda_{\min}(\nabla^2 \mathcal{F}(x)) \geq \frac{1}{5}\|u\|_2^2.$$

(p4) For any $x \in \mathcal{R}_3$, we have

$$\|\nabla \mathcal{F}(x)\|_2 > \frac{\|u\|_2^3}{8}.$$

The rank 1 setting is intuitive since there is only one strict saddle point and 2 isolated global minima. It is also important to notice that

$$\mathcal{R}_1 \cup \mathcal{R}_2 \cup \mathcal{R}_3 = \mathbb{R}^n.$$

Thus, the entire space $\mathbb{R}^n$ is parameterized by one of the regions: (I) the neighborhood of the strict saddle point, where the Hessian matrix $\nabla^2 \mathcal{F}(x)$ has negative eigenvalues; (II) the neighborhood of the global minima, where $\mathcal{F}(x)$ is strongly convex; and (III) the gradient $\nabla \mathcal{F}(x)$ has a sufficiently large norm. To better understand the landscape, we provide a visualization of the objective function $\mathcal{F}(x)$ in Figure 3 (a and b). We set $u = [1\ -1]^\top$, thus $M^* = uu^\top = \begin{bmatrix} 1 & -1 \\ -1 & 1 \end{bmatrix}$. It is easy to see that $x = [0\ 0]^\top$ is a strict saddle point and $x = \pm u$ are global minima, which matches with our analysis.

## 3.2 General Ranks

We then consider the general setting of $r \geq 1$, where $M^* = UU^\top$, $U \in \mathbb{R}^{n \times r}$. Characterizing the global landscape becomes much more involved as neither the strict saddle point nor the global minimum is isolated. Recall that we consider

$$\min_{X \in \mathbb{R}^{n \times r}} \mathcal{F}(X), \text{ where } \mathcal{F}(X) = \frac{1}{4}\|M^* - XX^\top\|_F^2. \tag{9}$$

For notational convenience, for any matrix $X$, we define:

$$\Psi_X \triangleq \arg\min_{\Psi \in \mathbb{O}_r} \|X - U\Psi\|_2 \text{ and } K_X \triangleq \begin{bmatrix} X_{(*,1)}X_{(*,1)}^\top & X_{(*,2)}X_{(*,1)}^\top & \cdots & X_{(*,r)}X_{(*,1)}^\top \\ X_{(*,1)}X_{(*,2)}^\top & X_{(*,2)}X_{(*,2)}^\top & \cdots & X_{(*,r)}X_{(*,2)}^\top \\ \vdots & \vdots & \ddots & \vdots \\ X_{(*,1)}X_{(*,r)}^\top & X_{(*,2)}X_{(*,r)}^\top & \cdots & X_{(*,r)}X_{(*,r)}^\top \end{bmatrix}. \tag{10}$$

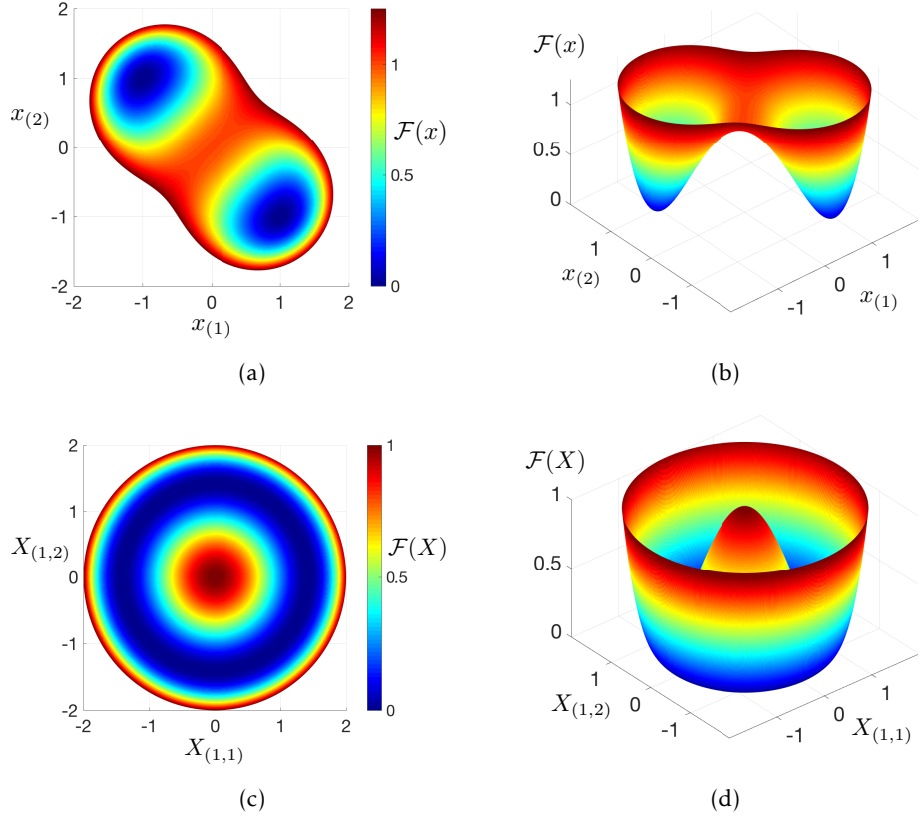

Figure 3: The visualization of objective functions $\mathcal{F}(X)$ for $r = 1$ (a and b) and $r = 2$ (c and d) using contour plots. In the case $r = 1$, the global minima are $x = [x_{(1)}\ x_{(2)}]^\top = [1\ -1]^\top$ and $[-1\ 1]^\top$. In the case $r = 2$, the global minima are $X = [X_{(1,1)}\ X_{(1,2)}]\Psi = [1\ -1]\Psi$ for all $\Psi \in \mathbb{O}_2$, i.e. any $X$ with $\|X\|_2 = \sqrt{2}$ is a global minimum. Note that we can only visualize $X \in \mathbb{R}^{1\times 2}$ when $r = 2$. Here $M^* = UU^\top = [1\ -1][1\ -1]^\top = 2$ is not low-rank in fact, and $X = [0\ 0]$ is not a strict saddle point but a local maximum.

Further, we introduce two sets:

$$\mathcal{X} = \left\{ X = \Phi\Sigma_2\Theta_2 \mid U \text{ has the SVD } U = \Phi\Sigma_1\Theta_1,\ (\Sigma_2^2 - \Sigma_1^2)\Sigma_2 = 0, \Theta_2 \in \mathbb{O}_r \right\} \text{ and}$$
$$\mathcal{U} = \{X \in \mathcal{X} \mid \Sigma_2 = \Sigma_1\}.$$

The set $\mathcal{X}$ contains all strict saddle points, and $\mathcal{U}$ is the set of all global minima, which will be proved in the following theorem. Specifically, for any $X$ that has a strict subset of the column bases of $U$ and identical corresponding singular values, $X$ is a strict saddle point of $\mathcal{F}$. This indicates that the strict saddle points are not isolated, and there are infinite many of them due to rotations (their measures in $\mathbb{R}^{n\times r}$ are zero). On the other hand, when $X$ is different from $U$ only by a rotation, $X$ is also a global minimum of $\mathcal{F}$.

By algebraic calculation, the gradient and the Hessian matrix of $\mathcal{F}(X)$, respectively, are

$$\nabla \mathcal{F}(X) = (XX^\top - M^*)X \in \mathbb{R}^{n\times r} \text{ and} \tag{11}$$

$$\nabla^2 \mathcal{F}(X) = K_X + I_r \otimes XX^\top + X^\top X \otimes I_n - I_r \otimes M^* \in \mathbb{R}^{rn\times rn}. \tag{12}$$

The gradient (11) and the Hessian matrix (12) for the general rank $r \geq 1$ reduce to (8) when $r = 1$. We provide the key arguments for the general rank setting in the following theorem. The proof is provided in Appendix C.

**Theorem 4.** Consider (9) for the general rank $r \geq 1$ and define the following regions:

$$\mathcal{R}_1 \triangleq \left\{ Y \in \mathbb{R}^{n\times r} \mid \sigma_r(Y) \leq \frac{1}{2}\sigma_r(U), \|YY^\top\|_F \leq 4\|UU^\top\|_F \right\},$$

$$\mathcal{R}_2 \triangleq \left\{ Y \in \mathbb{R}^{n\times r} \mid \min_{\Psi \in \mathcal{O}_r} \|Y - U\Psi\|_2 \leq \frac{\sigma_r^2(U)}{8\sigma_1(U)} \right\},$$

$$\mathcal{R}_3' \triangleq \left\{ Y \in \mathbb{R}^{n\times r} \mid \sigma_r(Y) > \frac{1}{2}\sigma_r(U), \min_{\Psi \in \mathcal{O}_r} \|Y - U\Psi\|_2 > \frac{\sigma_r^2(U)}{8\sigma_1(U)}, \|YY^\top\|_F \leq 4\|UU^\top\|_F \right\}, \text{ and}$$

$$\mathcal{R}_3'' \triangleq \left\{ Y \in \mathbb{R}^{n\times r} \mid \|YY^\top\|_F > 4\|UU^\top\|_F \right\}.$$

Then the following properties hold.

(p1) For any $X \in \mathcal{X}$, $X$ is a stationary point of $\mathcal{F}(X)$.

(p2) For any $X \in \mathcal{X}\setminus\mathcal{U}$, $X$ is a strict saddle point with $\lambda_{\min}(\nabla^2 \mathcal{F}(X)) \leq -\lambda_{\max}^2(\Sigma_1 - \Sigma_2)$. Moreover, for any $X \in \mathcal{R}_1$, $\nabla^2 \mathcal{F}(X)$ contains a negative eigenvalue, i.e.

$$\lambda_{\min}(\nabla^2 \mathcal{F}(X)) \leq -\frac{\sigma_r^2(U)}{4}.$$

(p3) For any $X \in \mathcal{U}$, $X$ is a global minimum of $\mathcal{F}(X)$, and $\nabla^2 \mathcal{F}(X)$ is positive semidefinite, which has $r(r-1)/2$ zero eigenvalues with the minimum nonzero eigenvalue at least $\sigma_r^2(U)$. Moreover, for any $X \in \mathcal{R}_2$, we have

$$z^\top \nabla^2 \mathcal{F}(X) z \geq \frac{1}{5}\sigma_r^2(U)\|z\|_2^2$$

for any $z \perp \mathcal{E}$, where $\mathcal{E} \subseteq \mathbb{R}^{n\times r}$ is a subspace spanned by all eigenvectors of $\nabla^2 \mathcal{F}(K_E)$ associated with negative eigenvalues, where $E = X - U\Psi_X$ and $\Psi_X$ and $K_E$ are defined in (10).

(p4) Further, we have

$$\|\nabla \mathcal{F}(X)\|_F > \frac{\sigma_r^4(U)}{9\sigma_1(U)} \text{ for any } X \in \mathcal{R}_3' \text{ and } \|\nabla \mathcal{F}(X)\|_F > \frac{3}{4}\sigma_1^3(X) \text{ for any } X \in \mathcal{R}_3''.$$

The following proposition shows that any $X \in \mathbb{R}^{n\times r}$ belongs to one of the four regions above. The proof is provided in Appendix G.1.

**Proposition 1.** Consider the four regions defined in Theorem 4, we have

$$\mathcal{R}_1 \cup \mathcal{R}_2 \cup \mathcal{R}_3' \cup \mathcal{R}_3'' = \mathbb{R}^{n \times r}.$$

Different from the rank 1 setting, we have one more region $\mathcal{R}_3''$, where the gradient has a sufficiently large norm. When $r = 1$, we have $\mathbb{O}_1 = \{1, -1\}$. Thus $\mathcal{X}$ reduces to $\{0\}$ and $\mathcal{U}$ reduces to $\{u, -u\}$, which matches with the result in Theorem 3. From (p2) of Theorem 4, we have that $X$ is approximately rank deficient in $\mathcal{R}_1$ since $\sigma_r(X) \leq \frac{1}{2}\sigma_r(U)$. From (p3) of Theorem 4, we have that $\mathcal{F}(X)$ is convex at a global minimum, rather than strongly convex. Moreover, in the neighborhood of a global minimum, $\mathcal{F}(X)$ is only strongly convex along certain directions. Analogous results are also provided in previous literature. For example, Tu et al. (2015) (in the analysis of Theorem 3.2) show that for any $X$ that satisfies $\|X - U\Psi_X\|_2 \leq c_1 \sigma_r(U)$, we have

**Regularity Property:** $\langle \nabla \mathcal{F}(X), X - U\Psi_X \rangle \geq c_2 \sigma_r^2(U) \|X - U\Psi_X\|_F^2 + c_3 \|UU^\top - XX^\top\|_F^2,$ (13)

where $c_1, c_2$, and $c_3$ are positive real constants. This indicates that when $X$ is close to a global minimum, $\mathcal{F}(X)$ is only strongly convex along the direction of $E = X - U\Psi_X$ (Procrustes difference). But our results are much more general. Specifically, we guarantee in (p3) of Theorem 4 that $\mathcal{F}(X)$ is strongly convex along all directions that are orthogonal to the subspace spanned by eigenvectors associated with negative eigenvalues of $\nabla^2 \mathcal{F}(K_E)$ for $K_E = X - U\Psi_X$. As we have shown in the analysis, there are at most $r(r-1)/2$ such directions potentially associated with the negative eigenvalues of $\nabla^2 \mathcal{F}(K_E)$. In other words, there are at least $nr - r(r-1)/2$ such directions, where $\mathcal{F}(X)$ is strongly convex. In the following lemma, we further show that $\mathcal{F}(X)$ is nonconvex in any neighborhood of a global minimum. The proof is provided in Appendix G.2.

**Proposition 2.** Let $\mathcal{B}_\varepsilon(U) = \{X \mid \|X - U\|_2 \leq \varepsilon\}$ be a neighborhood of $U$ with radius $\varepsilon > 0$. Then $\mathcal{F}(X)$ is a nonconvex function in $\mathcal{B}_\varepsilon(U)$.

We provide a visualization of the objective function $\mathcal{F}(X)$ in Figure 3 (c and d) by setting $r = 2$ and $U = [1 \ -1]$. The observation is that any $X$ satisfying $X = U\Psi_2$ is a global minimum, where $\Psi_2 \in \mathbb{O}_2$. Moreover, if we restrict $X$ to be a convex combination of any two distinct global minima, then $\mathcal{F}(X)$ is nonconvex, as we have shown in Proposition 2. Note that we can only visualize the case of $X \in \mathbb{R}^{1 \times 2}$, which results in a full rank $M^* = UU^\top = 2$ here. Thus $X = [0 \ 0]$ is a not strict saddle point in this degenerated example.

### 3.3 General Rectangular Matrices

We further discuss briefly on the scenario where the low-rank matrix is a general rectangular matrix. Recall that for $M^* = UV^\top \in \mathbb{R}^{n \times m}$ for some $U \in \mathbb{R}^{n \times r}$ and $V \in \mathbb{R}^{m \times r}$, we consider

$$\min_{X \in \mathbb{R}^{n \times r}, Y \in \mathbb{R}^{m \times r}} \mathcal{F}(X, Y), \text{ where } \mathcal{F}(X, Y) = \frac{1}{2}\|M^* - XY^\top\|_F^2. \quad (14)$$

Compared with the PSD matrix scenario (9) with $M^* \succeq 0$, it has one more issue of scaling invariance for the general rectangular matrix (14). Specifically, in addition to the rotation invariance as in the PSD case, when we multiply $X$ and divide $Y$ by an identical (nonzero) constant, $\mathcal{F}(X,Y)$ is also invariant. This results in a significantly increasing complexity of the structure for both strict saddle points and global minima. Moreover, the scaling issue also leads to a badly conditioned problem, e.g., when $\|X\|_F^2$ is very small and $\|Y\|_F^2$ is very large with $XY^\top$ fixed.

For ease of discussion, we provide an example when $n = m = r = 1$. Suppose $M^* = 1$, then the objective in (14) is $\mathcal{F}(x,y) = \frac{1}{2}(1 - xy)^2$. The corresponding Hessian matrix is

$$\nabla^2 \mathcal{F}(x,y) = \begin{bmatrix} y^2 & 2xy - 1 \\ 2xy - 1 & x^2 \end{bmatrix}.$$

It is easy to see that any $(x,y)$ satisfying $xy = 1$ is a global minimum, which makes the structure of the global minimum much more complicated than the PSD matrix case with rank $r = 1$ (only two global minima points in Figure 3). A visualization of $\mathcal{F}(x,y)$ is provided in Figure 4 (panel a and b). On the other hand, the problem becomes poorly conditioned, i.e., $\lambda_{\max}(\nabla^2 \mathcal{F}(x,y))/\lambda_{\min}(\nabla^2 \mathcal{F}(x,y)) \to \infty$ when $\|x\|_2 \to 0$ and $\|y\|_2 \to \infty$ with $xy = 1$.

To avoid such a scaling issue, we consider a regularized form as follows,

$$\min_{X \in \mathbb{R}^{n \times r}, Y \in \mathbb{R}^{m \times r}} \mathcal{F}_\lambda(X,Y), \text{ where } \mathcal{F}_\lambda(X,Y) = \frac{1}{2}\|M^* - XY^\top\|_F^2 + \frac{\lambda}{4}\|X^\top X - Y^\top Y\|_F^2. \quad (15)$$

where $\lambda > 0$ is a regularization parameter. Such a regularization has been considered in related problems of low-rank matrix factorization Tu et al. (2015); Yi et al. (2016), which enforces positive curvature when $X$ and $Y$ have similar spectrum to avoid the scaling issue discussed above.

Taking the example discussed above again, we have the regularized objective as $\mathcal{F}_\lambda(x,y) = \frac{1}{2}(1-xy)^2 + \frac{\lambda}{4}(x^2 - y^2)^2$ and the corresponding Hessian matrix as

$$\nabla^2 \mathcal{F}_\lambda(x,y) = \begin{bmatrix} (1-\lambda)y^2 + 3\lambda x^2 & 2(1-\lambda)xy - 1 \\ 2(1-\lambda)xy - 1 & (1-\lambda)x^2 + 3\lambda y^2 \end{bmatrix}.$$

With a proper value of $\lambda$, $\mathcal{F}_\lambda(x,y)$ has strong convexity in the neighborhood of $x = y = 1$ and $x = y = -1$, resulting in a much simplified structure of global minima, analogous to the PSD rank $r = 1$ case. A visualization of of $\mathcal{F}_\lambda(x,y)$ with $\lambda = 0.5$ is provided in Figure 4 (panel c and d). Compared with the objective $\mathcal{F}$ without a regularization, the regularized objective $\mathcal{F}_\lambda$ is much better conditioned even when one of $\|x\|_2$ and $\|y\|_2$ is very small and the other is very large.

We remark that after the initial release of our paper, Zhu et al. (2017) provide an extension of our analysis to the case of general rectangular matrices using the lifting formulation. Specifically, they show $U^\top U = V^\top V$ (Lemma 3 therein) at stationary points in the noiseless case, which implies that the stationary points are not affected by the regularization function in (15). Beyond stationary points, careful characterization is required to deal with the regularization, which is a fourth order polynomial on the factors (similar to the loss function). Consequently, they achieve analogous geometric result to our Theorem 4 for the asymmetric case.

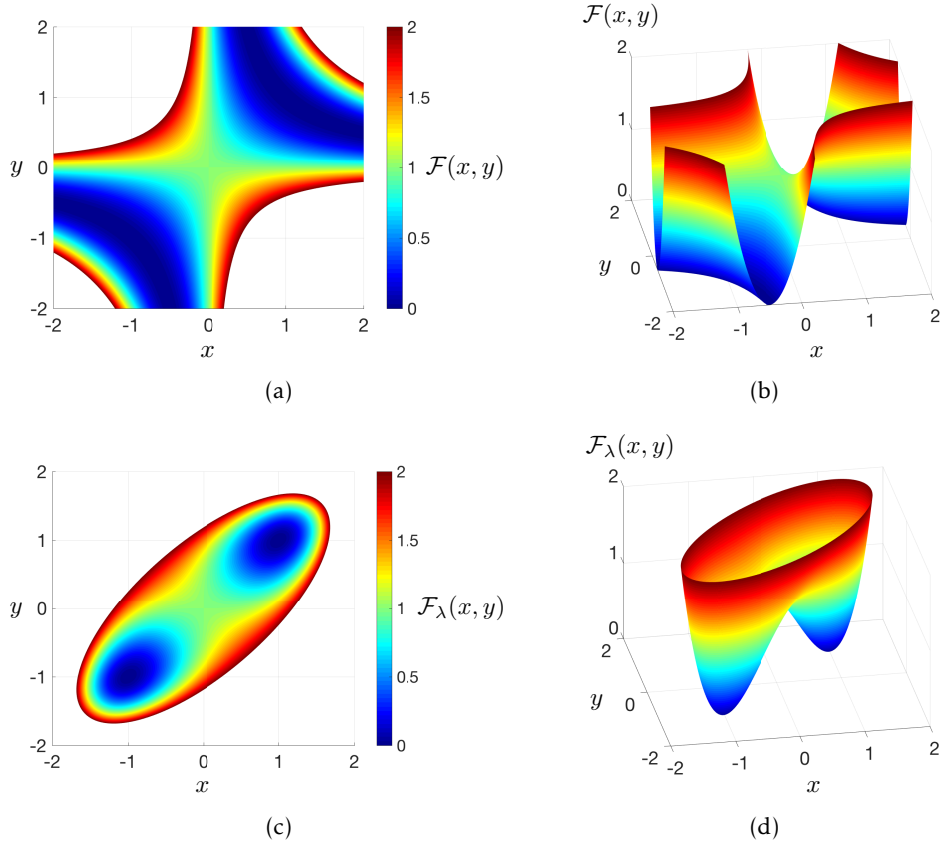

Figure 4: The visualization of objective functions $\mathcal{F}(x, y)$ with $u = v = 1$ (a, b) and $\mathcal{F}_\lambda(x, y)$ with $u = v = 1$ and $\lambda = 0.5$ (c, d). For $f(x, y)$, any $(x, y)$ that satisfies $xy = 1$ is a global minimum. For $\mathcal{F}_\lambda(x, y)$, $x = y = 1$ and $x = y = -1$ are the only global minima.

## 4 Matrix Sensing via Factorization

We extend our geometric analysis to the matrix sensing problem, which can be considered as a perturbed version of the low-rank matrix factorization problem. For simplicity, we first introduce the noiseless scenario and the noisy setting is discussed later, both of which preserve the entire landscape of optimization in the matrix factorization problem.

### 4.1 Matrix Sensing as a Perturbed Matrix Factorization Problem

We start with a formal description of the matrix sensing problem. For all $i \in [d]$, suppose $A_i \in \mathbb{R}^{n \times n}$ has i.i.d. zero mean sub-Gaussian entries with variance 1, then we observe

$$y_{(i)} = \langle A_i, M^* \rangle,$$

where $M^* \in \mathbb{R}^{n \times n}$ is a low-rank PSD matrix with $\text{Rank}(M^*) = r$. Denote $M^* = UU^\top$, where $U \in \mathbb{R}^{n \times r}$, then $y_{(i)} = \langle A_i, UU^\top \rangle$ and we recover $U$ by solving

$$\min_X F(X), \text{ where } F(X) = \frac{1}{4d} \sum_{i=1}^d \langle A_i, XX^\top - M^* \rangle^2. \tag{16}$$

The gradient and the Hessian matrix of $F(X)$, respectively, are

$$\nabla F(X) = \frac{1}{2d} \sum_{i=1}^d \langle A_i, XX^\top - M^* \rangle \cdot (A_i + A_i^\top) X \text{ and} \tag{17}$$

$$\nabla^2 F(X) = \frac{1}{2d} \sum_{i=1}^d I_r \otimes \langle A_i, XX^\top - M^* \rangle \cdot (A_i + A_i^\top) + \text{vec}\big((A_i + A_i^\top)X\big) \cdot \text{vec}\big((A_i + A_i^\top)X\big)^\top. \tag{18}$$

We first show the connection between the matrix sensing problem and the low-rank matrix factorization problem in the following lemma. The proof is provided in Appendix G.3.

**Lemma 1.** We have $\mathbb{E}(F(X)) = \mathcal{F}(X)$, $\mathbb{E}(\nabla F(X)) = \nabla \mathcal{F}(X)$, and $\mathbb{E}(\nabla^2 F(X)) = \nabla^2 \mathcal{F}(X)$.

From Lemma 1, we have that the objective (16), the gradient (17), and the Hessian matrix (18) of the matrix sensing problem are unbiased estimators of the counterparts of the low-rank matrix factorization problem in (9), (11), and (12) respectively. We then provide a finite sample perturbation bound for the gradient and the Hessian matrix of the matrix sensing problem. The proof is provided in Appendix G.4.

**Lemma 2.** Suppose $N \geq \max\{\|XX^\top - M^*\|_F^2, \|X\|_F^2, 1\}$. Given $\delta > 0$, if $d$ satisfies

$$d = \Omega(N \max\{nr, \sqrt{nr} \log(nr)\}/\delta),$$

then with high probability, we have

$$\|\nabla^2 F(X) - \nabla^2 \mathcal{F}(X)\|_2 \leq \delta \text{ and } \|\nabla F(X) - \nabla \mathcal{F}(X)\|_F \leq \delta.$$

From Lemma 2, we have that the landscape of the gradient and the Hessian matrix of low-rank matrix factorization is preserved for matrix sensing with high probability based on the concentrations of sub-Gaussian designs $\{A_i\}_{i=1}^d$, as long as the sample size $d$ is sufficiently large. These further allow us to derive the key properties (p1) – (p4) for matrix sensing directly from the counterparts of low-rank matrix factorization in Theorem 4. We formalize the result in the following Theorem. The proof is provided in Appendix D.

**Theorem 5.** Consider (16) for the general rank $r \geq 1$. If $d$ satisfies

$$d \geq C \cdot \max\left\{nr^2, n\sqrt{r} \log(n), r\sqrt{nr} \log(nr)\right\},$$

where $C > 0$ is a generic real constant, then with high probability, we have the following properties.

(p1) For any $X \in \mathcal{U} \cup \{0\}$, $X$ is a stationary point of $F(X)$.

(p2) $X = 0$ is a strict saddle point with $\lambda_{\min}(F(0)) \leq -\frac{7}{8}\|U\|_2^2$. Moreover, for any $X \in \mathcal{R}_1$, $\nabla^2 F(X)$ contains a negative eigenvalue, i.e.

$$\lambda_{\min}(\nabla^2 F(X)) \leq -\frac{\sigma_r^2(U)}{8}.$$

(p3) For any $X \in \mathcal{U}$, $X$ is a global minimum, and $\nabla^2 F(X)$ is positive semidefinite. Moreover, for any $X \in \mathcal{R}_2$, we have

$$z^\top \nabla^2 F(X) z \geq \frac{1}{10} \sigma_r^2(U) \|z\|_2^2$$

for any $z \perp \mathcal{E}$, where $\mathcal{E} \subseteq \mathbb{R}^{n \times r}$ is a subspace is spanned by all eigenvectors of $\nabla^2 \mathcal{F}(K_E)$ associated with negative eigenvalues, where $E = X - U\Psi_X$ and $\Psi_X$ and $K_E$ are defined in (10).

(p4) Further, we have

$$\|\nabla F(X)\|_F > \frac{\sigma_r^4(U)}{18\sigma_1(U)} \quad \text{for any } X \in \mathcal{R}_3' \quad \text{and} \quad \|\nabla F(X)\|_F > \frac{1}{4}\sigma_1^3(X) \quad \text{for any } X \in \mathcal{R}_3''.$$

From Theorem 5, we have that the landscape of the low-rank matrix factorization problem is preserved for the matrix sensing problem given a sufficiently large sample size $d$. This is to say, $F(X)$ has a negative curvature in the neighborhoods of strict saddle points, strong convexity along certain directions in the neighborhoods of global minima, and a sufficiently large norm for the gradient in the rest of domain. On the other hand, due to random perturbations by sensing matrices $\{A_i\}_{i=1}^d$, the set of strict saddle points in $\mathcal{X} \backslash \mathcal{U}$ reduces to $\{0\}$, while the rest of the points in $\mathcal{X} \backslash \mathcal{U}$ are nearly strict saddle.

## 4.2 Noisy Observation

We further consider a noisy scenario of the matrix sensing problem. Specifically, suppose $\{A_i\}_{i=1}^d$ are random matrices described above, then we observe

$$y_{(i)} = \langle A_i, M^* \rangle + z_{(i)} \text{ for all } i \in [d],$$

where $\{z_{(i)}\}_{i=1}^d$ are independent zero mean sub-Gaussian random noise with variance $\sigma_z^2$. Consequently, denoting $M^* = UU^\top$, we recover $U$ by solving

$$\min_X F(X), \text{ where } F(X) = \frac{1}{4d} \sum_{i=1}^d \left( \langle A_i, XX^\top - M^* \rangle - z_{(i)} \right)^2. \tag{19}$$

We then provide the key properties (p1) – (p4) for the noisy version of the matrix sensing problem in the following corollary. The proof is provided in Appendix E.

**Corollary 2.** Consider (19) for the general rank $r \geq 1$. Given $\varepsilon > 0$, if $d$ satisfies

$$d \geq \frac{C\sigma_z^2 \cdot \max\left\{nr^2,\, n\sqrt{r}\log(n),\, r\sqrt{nr}\log(nr)\right\}}{\varepsilon^2},$$

where $C > 0$ is a generic real constant, then with high probability, we have that properties (p1) – (p4) in Theorem 5 hold, as well as the following estimation error

$$\|\widehat{M} - M^*\|_F^2 = \mathcal{O}(\varepsilon^2),$$

where $\widehat{M} = \widehat{X}\widehat{X}^\top$ for $\widehat{X} = \arg\min_X F(X)$ in (19).

Compared with Theorem 5, the sufficient sample complexity for preserving the key properties (p1) – (p4) of the landscape in Corollary 2 has one more dependence on the variance of noise, which is a natural result for noisy measurements. We remark that preserving the global landscape is more challenging than guaranteeing the convergence to a local minimum within the optimal distance to the true model parameter, which only requires a local analysis in a neighborhood of the true model parameter. Existing results only discuss some local geometry instead of the global one as we do, such as the strict saddle points and the neighborhood of true model parameter Chen and Wainwright (2015); Bhojanapalli et al. (2016b).

## 5 Discussion

We provide further discussion on extending our analysis for matrix sensing to achieve the optimal sample complexity by relaxing the geometric properties as a tradeoff. In addition, we make some comments on how the geometric analysis in this paper can imply strong convergence guarantees for several popular iterative algorithms.

### 5.1 From Suboptimal to Optimal Sampling Complexity for Matrix Sensing

The sampling complexity is $\mathcal{O}(nr^2)$ for matrix sensing when we preserve the entire landscape of the matrix factorization problem (9). If we relax the properties of optimization landscape to be preserved, the optimal complexity $\mathcal{O}(nr)$ can be attained. In specific, consider the noiseless scenario by solving (16). Then we have the following geometric properties for matrix sensing. The proof is provided in Appendix F.

**Theorem 6.** Consider (9) for the general rank $r \geq 1$ and define the following regions:

$$\mathcal{R}_1 \triangleq \left\{ Y \in \mathbb{R}^{n \times r} \mid \min_{\Psi \in \mathcal{O}_r} \|Y - U\Psi\|_2 > \frac{\sigma_r(U)}{4}, \|\nabla F(Y)\|_F \leq \frac{\sigma_r^3(U)}{96} \right\},$$

$$\mathcal{R}_2 \triangleq \left\{ Y \in \mathbb{R}^{n \times r} \mid \min_{\Psi \in \mathcal{O}_r} \|Y - U\Psi\|_2 \leq \frac{\sigma_r(U)}{4} \right\}, \text{ and}$$

$$\mathcal{R}_3 \triangleq \left\{ Y \in \mathbb{R}^{n \times r} \mid \min_{\Psi \in \mathcal{O}_r} \|Y - U\Psi\|_2 > \frac{\sigma_r(U)}{4}, \|\nabla F(Y)\|_F > \frac{\sigma_r^3(U)}{96} \right\},$$

If $d$ satisfies

$$d \geq C \cdot nr,$$

where $C$ is a generic real constant, then with high probability, we have the following properties.

(p1) For any $X \in \mathcal{U} \cup \{0\}$, $X$ is a stationary point of $F(X)$.

(p2) [Direct result from Ge et al. (2017)] For any $X \in \mathcal{R}_1$, including the strict saddle point $X = 0$, $\nabla^2 F(X)$ contains a negative eigenvalue, i.e.

$$\lambda_{\min}(\nabla^2 F(X)) \leq -\frac{1}{6}\sigma_r^2(U).$$

(p3) For any $X \in \mathcal{U}$, $X$ is a global minimum, and $\nabla^2 F(X)$ is positive semidefinite. Moreover, for any $X \in \mathcal{R}_2$ with $\Psi_X$ defined in (10), we have

$$\langle \nabla F(X), X - U\Psi_X \rangle \geq \frac{\sigma_r^2(U)}{4}\|X - U\Psi_X\|_F^2 + \frac{1}{20\|U\|_2}\|\nabla F(X)\|_F^2.$$

(p4) Further, for any $X \in \mathcal{R}_3$, we have

$$\|\nabla F(X)\|_F > \frac{\sigma_r^3(U)}{96}.$$

It is immediate from Theorem 6 that we have

$$\mathcal{R}_1 \cup \mathcal{R}_2 \cup \mathcal{R}_3 = \mathbb{R}^n.$$

When $d = \Omega(nr)$, weaker properties of optimization landscape can be obtained. First of all, unlike $\mathcal{R}_1$ of Theorem 4, it is not clear whether there is (approximate) rank deficiency in $\mathcal{R}_1$ from Theorem 6. Since the rank deficiency is a key reason for generating strict saddle points, we face a gap in the geometric interpretation. Moreover, in the neighborhood of global minima in (p3), we have the regularity property (13). As we have discussed after Theorem 4, this is a weaker result than (p3) therein, which can guarantee the strong convexity in a larger number of directions. We suspect that this is a tradeoff between the optimal sample complexity and strong geometric properties (though this may be a proof artifact). In addition, the characterization of both regions $\mathcal{R}_1$ and $\mathcal{R}_3$ in Theorem 6 depend on both problem parameter $X$ and sensing matrices $\{A_i\}_{i=1}^d$ (embedded in $\nabla F(X)$). This makes the regions less explicit than $\mathcal{R}_1$ and $\mathcal{R}_3$ in Theorem 5, which only on $X$.

We further address a brief comparison with Tu et al. (2015); Ge et al. (2017). Our Theorem 6 has slightly stronger geometric guarantees than Tu et al. (2015); Ge et al. (2017) under the same conditions. In specific, due to a refined analysis, our neighborhood of global minima $\mathcal{R}_2$ characterized via the spectral norm of the Procrustes difference is larger than the corresponding region in Tu et al. (2015); Ge et al. (2017) characterized via the Frobenius norm, i.e.,

$$\left\{Y \in \mathbb{R}^{n \times r} \mid \min_{\Psi \in \mathcal{O}_r} \|Y - U\Psi\|_F \leq \frac{\sigma_r(U)}{4}\right\} \subset \left\{Y \in \mathbb{R}^{n \times r} \mid \min_{\Psi \in \mathcal{O}_r} \|Y - U\Psi\|_2 \leq \frac{\sigma_r(U)}{4}\right\} \text{ for all } \mathrm{rank}(U) > 1.$$

Moreover, Tu et al. (2015) only provide a local geometric property in the neighborhood of global minima $\mathcal{R}_2$ using the regularity property. In contrast, we provide a global one in Theorem 6.

## 5.2 Convergence of Iterative Algorithms

Here are some comments on the convergence guarantees. With the explicit geometry of the objective function, it is straightforward to provide convergence guarantees using many popular iterative algorithms, even without special initializations. A few examples of recent progress on related nonconvex problems are listed as follows.

- A trust-region type of algorithm is proposed in Sun et al. (2016) to solve a specific type of nonconvex problem, i.,e., phase retrieval. Similar to our analysis, the authors explicitly divide the whole domain into three overlapping regions $\mathcal{R}_1$, $\mathcal{R}_2$, and $\mathcal{R}_3$, based on which they show a sufficient decrease of objective in $\mathcal{R}_1$ and $\mathcal{R}_3$ and an overall R-quadratic convergence to a global minimum. Another closely related algorithm is the second-order majorization type of algorithm proposed in Carmon and Duchi (2016), which finds an $\varepsilon$-second-order stationary point $x_\varepsilon$ for a predefined precision $\varepsilon > 0$, i.e.,

$$\|\nabla f(x_\varepsilon)\|_2 \leq \varepsilon \text{ and } \nabla^2 f(x_\varepsilon) \succeq -\sqrt{\beta\varepsilon}I$$

for general lower bounded objective $f$ that has a Lipschitz gradient and a $2\beta$-Lipschitz Hessian. The algorithm is based on iteratively solving a cubic-regularized quadratic approximation of the objective function using gradient descent steps, and an overall sublinear convergence guarantee is provided.

- A gradient descent algorithm is analyzed in Lee et al. (2016); Panageas and Piliouras (2016) for twice-continuously differentiable functions with a Lipschitz gradient. The authors provide an asymptotic convergence guarantee of Q-linear convergence to a local minimum if all saddle points are strict saddle.

- A noisy stochastic gradient descent algorithm is proposed in Ge et al. (2015) for so-called strict saddle problems, i.e., any point the given objective function is in $\mathcal{R}_1$ (negative curvature in neighborhood of strict saddle points), $\mathcal{R}_3$ (the gradient has a sufficiently large norm), or a strongly convex neighborhood containing a local minimum. The authors show a sufficient decrease of objective for each noisy stochastic gradient step in $\mathcal{R}_1$ and $\mathcal{R}_3$, and an overall R-sublinear convergence to a local minimum.

The algorithms discussed above can be extended to solve the matrix factorization type of problems considered in this paper, with convergence guarantees. Note that for those requiring a local strong convexity, such as Ge et al. (2015), the analysis does not apply directly here for the matrix factorization type of problems in general. This can be settled by applying the Polyak-Lojasiewicz condition instead Polyak (1963); Karimi et al. (2016).

## 5.3 Extension to Matrix Completion

Finally, we comment on a closely related problem – matrix completion, where we expect similar global geometric properties to hold. Specifically, given a entry-wise observed matrix $\mathcal{P}_\Omega(M^*) \in \mathbb{R}^{n \times n}$ for $M^* \succeq 0$, where $\mathcal{P}_\Omega(M^*_{i,j}) = 0$ if $(i,j) \notin \Omega$ and $\mathcal{P}_\Omega(M^*_{i,j}) = M^*_{i,j}$ if $(i,j) \in \Omega$ for some subset $\Omega \subseteq [n] \times [n]$, we solve

$$\min_{X \in \mathbb{R}^{n \times r}} H(X) + R(X), \text{ where } H(X) = \frac{1}{p} \|\mathcal{P}_\Omega(M^* - XX^\top)\|_F^2. \tag{20}$$

where $p = |\Omega|/n^2$ is the sampling rate and $R(X)$ is a regularization function to enforce low coherence of $X$ (see more details in Sun and Luo (2015); Ge et al. (2016)). Similar to the matrix sensing problem, (20) can be also considered as a perturbed version of the low-rank matrix factorization problem (4). It is easy to see that if $\Omega$ is uniformly sampled over all subsets of $[n] \times [n]$ for a given cardinality, then we have

$$\mathbb{E}(H(X)) = \|M^* - XX^\top\|_F^2.$$

However, because the entry-wise sampling model is more challenging than the random linear measurement model and the incoherence of the low-rank matrix is generally required, the extra regularization term is inevitable for the matrix completion problem. This leads to a much more involved perturbation analysis for (20) than that of matrix sensing. For example, Sun and Luo (2015) establish the geometric analysis around the global minimizers; Ge et al. (2016) show that there exists no spurious local optima.

BIRNBAUM, A., JOHNSTONE, I. M., NADLER, B. and PAUL, D. (2013). Minimax bounds for sparse PCA with noisy high-dimensional data. *The Annals of Statistics*, **41** 1055.

BLONDEL, M., ISHIHATA, M., FUJINO, A. and UEDA, N. (2016). Polynomial networks and factorization machines: New insights and efficient training algorithms. *arXiv preprint arXiv:1607.08810*.

BURER, S. and MONTEIRO, R. D. (2003). A nonlinear programming algorithm for solving semidefinite programs via low-rank factorization. *Mathematical Programming*, **95** 329–357.

BURER, S. and MONTEIRO, R. D. (2005). Local minima and convergence in low-rank semidefinite programming. *Mathematical Programming*, **103** 427–444.

CAI, J.-F., CANDÈS, E. J. and SHEN, Z. (2010). A singular value thresholding algorithm for matrix completion. *SIAM Journal on Optimization*, **20** 1956–1982.

CAI, T. T., MA, Z., WU, Y. ET AL. (2013). Sparse PCA: Optimal rates and adaptive estimation. *The Annals of Statistics*, **41** 3074–3110.

CANDES, E. J., LI, X. and SOLTANOLKOTABI, M. (2015). Phase retrieval via wirtinger flow: Theory and algorithms. *IEEE Transactions on Information Theory*, **61** 1985–2007.

CANDES, E. J. and PLAN, Y. (2011). Tight oracle inequalities for low-rank matrix recovery from a minimal number of noisy random measurements. *IEEE Transactions on Information Theory*, **57** 2342–2359.

CANDÈS, E. J. and RECHT, B. (2009). Exact matrix completion via convex optimization. *Foundations of Computational Mathematics*, **9** 717–772.

CARMON, Y. and DUCHI, J. C. (2016). Gradient descent efficiently finds the cubic-regularized nonconvex newton step. *arXiv preprint arXiv:1612.00547*.

CHEN, Y. and WAINWRIGHT, M. J. (2015). Fast low-rank estimation by projected gradient descent: General statistical and algorithmic guarantees. *arXiv preprint arXiv:1509.03025*.

DUMMIT, D. S. and FOOTE, R. M. (2004). *Abstract algebra*, vol. 3. Wiley Hoboken.

FOUCART, S. and RAUHUT, H. (2013). *A mathematical introduction to compressive sensing*, vol. 1. Birkhäuser Basel.

GE, R., HUANG, F., JIN, C. and YUAN, Y. (2015). Escaping from saddle points—online stochastic gradient for tensor decomposition. In *Proceedings of The 28th Conference on Learning Theory*.

GE, R., JIN, C. and ZHENG, Y. (2017). No spurious local minima in nonconvex low rank problems: A unified geometric analysis. *arXiv preprint arXiv:1704.00708*.

GE, R., LEE, J. D. and MA, T. (2016). Matrix completion has no spurious local minimum. *arXiv preprint arXiv:1605.07272*.25

# A Proofs of Results in Section 2

## A.1 Proof of Theorem 1

From the directional derivative of $f$ at $x_{\mathcal{G}}$, for any $x$, we have

$$0 = \lim_{t \to 0} \frac{f(x_{\mathcal{G}} + tg(x)) - f(x_{\mathcal{G}} + tx)}{t} = \nabla f(x_{\mathcal{G}})^\top (g(x) - x),$$

which implies $x_{\mathcal{G}}$ is a stationary point.

## A.2 Proof of Theorem 2

Given any $v \in T_x G(x)$, there exists a smooth path $\gamma : (-1, 1) \to G(x)$ with $\gamma(0) = x$ and $v = \gamma'(0)$. We consider the function $\ell(t) = f(\gamma(t))$. By chain rule, we have

$$\ell'(t) = \nabla f(\gamma(t))^\top \gamma'(t) \text{ and } \ell''(t) = \gamma'(t)^\top \nabla^2 f(\gamma(t)) \gamma'(t) + \nabla f(\gamma(t))^\top \gamma''(t). \tag{21}$$

Furthermore, since $G$ is the invariant group, we have $\ell(t) = f(\gamma(t)) = \text{const}$ and $\ell'(t) = \ell''(t) = 0$ for any $t \in (-1, 1)$. Since $x$ is stationary, $\nabla f(\gamma(0)) = \nabla f(x) = 0$ and we plug it into (21) to have

$$0 = \ell''(0) = \gamma'(0)^\top \nabla^2 f(\gamma(0)) \gamma'(0) = v^\top H_x v,$$

which implies that $v \in \text{Null}(H_x)$. This completes our proof.

# B Proof of Theorem 3

We separate the analysis into four intermediate components, one for each claim. We first identifies the stationary point of $\mathcal{F}(x)$ in the following lemma. The proof is provided in Appendix B.1.

**Lemma 3.** $0$, $u$ and $-u$ are the only stationary points of $\mathcal{F}(x)$, i.e., $\nabla \mathcal{F}(x) = 0$.

Next, we characterize two types of stationary points. We show a stronger result in the following lemma that $x = 0$ is a strict saddle point, and $\nabla^2 \mathcal{F}(x)$ has both positive and negative eigenvalue in the neighborhood of $x = 0$. The proof is provided in Appendix B.2.

**Lemma 4.** $x = 0$ is a strict saddle point, where $\nabla^2 \mathcal{F}(0)$ is negative semi-definite with $\lambda_{\min}(\mathcal{F}(0)) = -\|u\|_2^2$. Moreover, for any $x \in \mathcal{R}_1$, $\nabla^2 \mathcal{F}(x)$ contains positive eigenvalues and negative eigenvalues, i.e.

$$\lambda_{\max}(\nabla^2 \mathcal{F}(x)) \geq \|x\|_2^2 \quad \text{and} \quad \lambda_{\min}(\nabla^2 \mathcal{F}(x)) \leq -\frac{1}{2}\|u\|_2^2.$$

Moreover, we identify that $x = \pm u$ are global minima, and $\mathcal{F}(x)$ is strongly convex in a neighborhood of $x = \pm u$. The proof is provided in Appendix B.3.

**Lemma 5.** For $x = \pm u$, $x$ is a global minimum, and $\nabla^2 \mathcal{F}(x)$ is positive definite with $\lambda_{\min}(\nabla^2 \mathcal{F}(x)) = \|u\|_2^2$. Moreover, for any $x \in \mathcal{R}_2$, $\mathcal{F}(x)$ is locally strongly convex, i.e.

$$\lambda_{\min}(\nabla^2 \mathcal{F}(x)) \geq \frac{1}{5}\|u\|_2^2.$$

Finally, we show that outside the regions $\mathcal{R}_1$ and $\mathcal{R}_2$, the gradient $\nabla \mathcal{F}(s)$ has a sufficiently large norm. The proof is provided in Appendix B.4.

**Lemma 6.** For any $x \in \mathcal{R}_3$, we have

$$\|\nabla \mathcal{F}(x)\|_2 > \frac{\|u\|_2^3}{8}.$$

Combining Lemma 3 – Lemma 6, we finish the proof.

## B.1 Proof of Lemma 3

We provide an algebraic approach to determine stationary points here. Without loss of generality, we assume $\|u\|_2 = 1$. Then we write

$$x = \alpha u + w,$$

where $\alpha \in \mathbb{R}$ is a constant and $w^\top u = 0$. Accordingly, we solve

$$(xx^\top - uu^\top)x = [(\alpha u + w)(\alpha u + w)^\top - uu^\top](\alpha u + w) = [\alpha^2 uu^\top + \alpha wu^\top + \alpha uw^\top + ww^\top - uu^\top](\alpha u + w)$$
$$= \alpha^3 u + \alpha^2 w + \alpha u\|w\|_2^2 + w\|w\|_2^2 - \alpha u = u(\alpha^3 + \alpha\|w\|_2^2 - \alpha) + w(\alpha^2 + \|w\|_2^2) = 0.$$

1. Suppose $\alpha = 0$, which implies $u^\top x = 0$. Thus we must have $(xx^\top - uu^\top)x = x\|x\|_2^2 = 0$, which further implies $x = 0$ is a stationary point.

2. Suppose $\|w\|_2 = 0$, which implies $w = 0$. Thus we must have $(xx^\top - uu^\top)x = (\alpha^3 - \alpha)u = 0$, which further implies $\alpha = -1$ or $1$, i.e., $x = -u$ and $x = u$ are stationary points.

3. Suppose $\alpha \neq 0$ and $w \neq 0$. We then require

$$\alpha^2 + \|w\|_2^2 - 1 = 0 \quad \text{and} \quad \alpha^2 + \|w\|_2^2 = 0.$$

This conflict with each other, which implies there is no stationary point when $\alpha \neq 0$ and $w \neq 0$.

The results are identical to those by applying generic theories in Section 2 directly.

## B.2 Proof of Lemma 4

We first show that $x = 0$ is a strict saddle point, by verifying that $\lambda_{\min}(\nabla^2 \mathcal{F}(0)) < 0$ and for any neighborhood $\mathcal{B}$ of $x = 0$, there exist $y_1, y_2 \in \mathcal{B}$ such that $\mathcal{F}(y_1) \leq \mathcal{F}(0) \leq \mathcal{F}(y_2)$.

From (8) we have $\nabla^2 \mathcal{F}(0) = -uu^\top$. For any $z \in \mathbb{R}^n$ with $\|z\|_2 = 1$, we have

$$z^\top \nabla^2 \mathcal{F}(0) z = -(z^\top u)^2 \geq -\|u\|_2^2,$$

where the last inequality is from Cauchy-Schwarz. Then we have $\nabla^2 \mathcal{F}(0)$ is negative semi-definite. The minimal eigenvalue is $\lambda_{\min}(\nabla^2 \mathcal{F}(0)) = -\|u\|_2^2$ with the corresponding eigenvector $u/\|u\|_2$ and the maximal eigenvalue is $\lambda_{\max}(\nabla^2 \mathcal{F}(0)) = 0$ with the corresponding eigenvector $z$ that satisfies $u^\top z = 0$.

Let $y_1 = \alpha u$, where $\alpha \in [0, 1]$, and $y_2$ be any vector that satisfies $y_2^\top u = 0$. Then we have

$$\mathcal{F}(y_1) = \frac{1}{4}\|uu^\top - \alpha^2 uu^\top\|_2^2 = \frac{(1-\alpha^2)}{4}\|uu^\top\|_2^2 \leq \frac{1}{4}\|uu^\top\|_2^2 = \mathcal{F}(0) \text{ and}$$

$$\mathcal{F}(y_2) = \frac{1}{4}\left(\|uu^\top\|_2^2 + \|y_2\|_2^2\right) \geq \mathcal{F}(0).$$

Therefore, we have $\mathcal{F}(y_1) \leq \mathcal{F}(0) \leq \mathcal{F}(y_2)$, which implies $x = 0$ is a strict saddle point.

Next, we show that for any $\|x\|_2 \leq \frac{1}{2}\|u\|_2$, $\nabla^2 \mathcal{F}(x)$ has both positive and negative eigenvalues. Given a point $x$, let $z_{\max}(x)$ and $z_{\min}(x)$ denote the eigenvectors of $\lambda_{\max}(\nabla^2 \mathcal{F}(x))$ corresponding to the largest and smallest eigenvalues respectively. Then for any $x \in \mathcal{R}_1$, $\nabla^2 \mathcal{F}(x)$ has at least a positive eigenvalue since

$$z_{\max}^\top(x) \nabla^2 \mathcal{F}(x) z_{\max}(x) \geq z_{\max}^\top(0) \nabla^2 \mathcal{F}(x) z_{\max}(0) = 2(z_{\max}^\top(0) x)^2 + \|x\|_2^2 \geq \|x\|_2^2.$$

On the other hand, we have $z_{\min}(0) = u/\|u\|_2$ and $\lambda_{\min}(\nabla^2 \mathcal{F}(0)) = -\|u\|_2^2$ from the previous discussion. Then for any $x \in \mathcal{R}_1$, $\nabla^2 \mathcal{F}(x)$ has at least a negative eigenvalue since

$$z_{\min}^\top(x) \nabla^2 \mathcal{F}(x) z_{\min}(x) \leq z_{\min}^\top(0) \nabla^2 \mathcal{F}(x) z_{\min}(0) = 2(z_{\min}^\top(0) x)^2 + \|x\|_2^2 - \|u\|_2^2$$

$$\leq 3\|x\|_2^2 - \|u\|_2^2 \leq -\frac{1}{4}\|u\|_2^2.$$

## B.3 Proof of Lemma 5

We only discuss the scenario when $x = u$. The argument for $x = -u$ is similar. From the Hessian matrix $\nabla^2 \mathcal{F}(x)$ in (8), we have $\nabla^2 \mathcal{F}(u) = uu^\top + \|u\|_2^2 \cdot I_n$. For any $z \in \mathbb{R}^n$ with $\|z\|_2 = 1$, we have

$$z^\top \nabla^2 \mathcal{F}(u) z = (z^\top u)^2 + \|u\|_2^2 \geq \|u\|^2,$$

then $\lambda_{\min}(\nabla^2 \mathcal{F}(u)) = \|u\|_2^2$ with the corresponding eigenvector $z$ satisfying $u^\top z = 0$. Therefore, $\nabla^2 \mathcal{F}(u)$ is positive definite and $x = u$ is a local minimum of $\mathcal{F}(x)$. Moreover, $x = u$ is a also a global minimum since

$$\mathcal{F}(u) = \min_{x \in \mathbb{R}^n} \mathcal{F}(x) = 0.$$

On the other hand, let $x = u + e$. For any $x \in \mathcal{R}_2$, we have

$$\left|z^\top \left(\nabla^2 \mathcal{F}(x) - \nabla^2 \mathcal{F}(u)\right)z\right| = \left|z^\top \left(2(u+e)(u+e)^\top + \|x\|_2^2 \cdot I_n - 2uu^\top - \|u\|_2^2 \cdot I_n\right)z\right|$$
$$= \left|z^\top \left(2(ee^\top + eu^\top + ue^\top) + (\|e\|_2^2 + 2e^\top u) \cdot I_n\right)z\right|$$
$$\leq \left(3\|e\|_2^2 + 6\|e\|_2\|u\|_2\right) \cdot \|z\|_2^2 \leq \frac{51}{64}\|u\|_2^2,$$

which further implies

$$z^\top \nabla^2 \mathcal{F}(x) z \geq z^\top \nabla^2 \mathcal{F}(u) z - \left|z^\top \left(\nabla^2 \mathcal{F}(x) - \nabla^2 \mathcal{F}(u)\right)z\right| \geq \frac{1}{5}\|u\|_2^2.$$

## B.4 Proof of Lemma 6

Let $x = \alpha u + \beta w \|u\|_2$, where $\alpha, \beta \in \mathbb{R}$, $w^\top u = 0$ and $\|w\|_2 = 1$. Then we have

$$\|\nabla \mathcal{F}(x)\|_2^2 = \|(xx^\top - uu^\top)x\|_2^2 = \|(\alpha^2 + \beta^2)\|u\|_2^2 \cdot (\alpha u + \beta w \|u\|_2) - \alpha \|u\|_2^2 \cdot u\|_2^2$$
$$= \|(\alpha^3 + \alpha\beta^2 - \alpha)\|u\|_2^2 \cdot u + \beta(\alpha^2 + \beta^2)\|u\|_2^3 \cdot w\|_2^2 = [(\alpha^3 + \alpha\beta^2 - \alpha)^2 + \beta^2(\alpha^2 + \beta^2)^2]\|u\|_2^6$$

Then region $\mathcal{R}_3$ is equivalent to the following set

$$\mathcal{X}_u = \left\{x = \alpha u + \beta w \|u\|_2 \mid \alpha^2 + \beta^2 > \frac{1}{4}, (\alpha - 1)^2 + \beta^2 > \frac{1}{64}\right\}.$$

By direct calculation, the infimum of $\|\nabla \mathcal{F}(x)\|_2$ subject to $x \in \mathcal{X}_u$ is obtained when $\alpha \to 0$ and $\beta \to \frac{1}{2}$, i.e., $\|\nabla \mathcal{F}(x)\|_2 > \frac{\|u\|_2^3}{8}$.

# C Proof of Theorem 4

The proof scheme is identical to that of the rank 1 case in Theorem 3. However, the analysis is much more challenging due to the nonisolated strict saddle points and minimum points.

First, we identify the stationary points of $\mathcal{F}(X)$ in the following lemma. The proof is provided in Appendix C.1.

**Lemma 7.** For any $X \in \mathcal{X}$, $X$ is a stationary point of $\mathcal{F}(X)$.

Next, we characterize two types of stationary points. We show a stronger result in the following lemma that for any $X \in \mathcal{X}$, it is a strict saddle point, where the Hessian matrix has both positive and negative eigenvalues. Further, the Hessian matrix has a negative eigenvalue in the neighborhood of $X \in \mathcal{X}$. The proof is provided in Appendix C.2.

**Lemma 8.** For any $X \in \mathcal{X} \backslash \mathcal{U}$, $X$ is a strict saddle point with

$$\lambda_{\min}(\nabla^2 \mathcal{F}(X)) \leq -\lambda_{\max}^2(\Sigma_1 - \Sigma_2) \text{ and } \lambda_{\max}(\nabla^2 \mathcal{F}(X)) \geq 2\lambda_{\max}^2(\Sigma_2).$$

Moreover, for any $X \in \mathcal{R}_1$, $\nabla^2 \mathcal{F}(X)$ contains a negative eigenvalue, i.e.

$$\lambda_{\min}(\nabla^2 \mathcal{F}(X)) \leq -\frac{\sigma_r^2(U)}{4}.$$

Moreover, we show in the following lemma that for any $X \in \mathcal{U}$, it is a global minimum, and $\mathcal{F}(X)$ is only strongly convex along certain directions in the neighborhood of $X \in \mathcal{U}$. The proof is provided in Appendix C.3.

**Lemma 9.** *For any $X \in \mathcal{U}$, $X$ is a global minimum of $\mathcal{F}(X)$, and $\nabla^2 \mathcal{F}(X)$ is positive semidefinite, which has exactly $r(r-1)/2$ zero eigenvalues with the minimum nonzero eigenvalue at least $\sigma_r^2(U)$. Moreover, for any $X \in \mathcal{R}_2$, we have*

$$z^\top \nabla^2 \mathcal{F}(X) z \geq \frac{1}{5} \sigma_r^2(U) \|z\|_2^2$$

*for any $z \perp \mathcal{E}$, where $\mathcal{E} \subseteq \mathbb{R}^{n \times r}$ is a subspace is spanned by all eigenvectors of $\nabla^2 \mathcal{F}(K_E)$ associated with the negative eigenvalues, where $E = X - U\Psi_X$ and $\Psi_X$ and $K_E$ are defined in (10).*

Finally, we show in the following lemma that the gradient $\nabla \mathcal{F}(X)$ has a sufficiently large norm outside the neighborhood of stationary points. The proof is provided in Appendix C.4

**Lemma 10.** *The gradient $\nabla \mathcal{F}(X)$ has sufficiently large norm in $\mathcal{R}_3'$ and $\mathcal{R}_3''$, i.e.,*

$$\|\nabla \mathcal{F}(X)\|_F > \frac{\sigma_r^4(U)}{9\sigma_1(U)} \quad \text{for any } X \in \mathcal{R}_3', \text{ and}$$

$$\|\nabla \mathcal{F}(X)\|_F > \frac{3}{4} \sigma_1^3(X) \quad \text{for any } X \in \mathcal{R}_3''.$$

Combining Lemma 7 – Lemma 10, we finish the proof.

## C.1 Proof of Lemma 7

We provide an algebraic approach to determine stationary points here. We denote $X = \Phi \Sigma_2 \Theta_2 + W$, where $W^\top \Phi = 0$. Accordingly, we solve

$$\begin{aligned}(XX^\top - UU^\top)X &= [(\Phi \Sigma_2 \Theta_2 + W)(\Phi \Sigma_2 \Theta_2 + W)^\top - UU^\top](\Phi \Sigma_2 \Theta_2 + W) \\ &= \Phi \Sigma_2 \Theta_2^\top (\Theta_2 \Sigma_2^2 \Theta_2 + W^\top W) + W(\Theta_2 \Sigma_2^2 \Theta_2 + W^\top W) - \Phi \Sigma_1^2 \Sigma_2 \Theta_2^\top = 0.\end{aligned}$$

1. Suppose $\Sigma_2 = 0$, which implies

$$WW^\top W = 0.$$

The solution to the equation above is $W = 0$, which indicates that $X = 0$ is a stationary point.

2. Suppose $W = 0$, which implies

$$\Phi(\Sigma_2^2 - \Sigma_1^2)\Sigma_2 \Theta_2^\top = 0.$$

The solution to the equation above is $(\Sigma_2^2 - \Sigma_1^2)\Sigma_2$, which indicates that $\Phi \Sigma_2 \Theta_2$ is a stationary point for any $\Theta_2 \in \mathcal{O}_r$ and $\Sigma_2 = \Sigma_1 I_{\text{Mask}}$, where $I_{\text{Mask}}$ is setting arbitrary number of diagonal elements of the identity matrix as $0$ at arbitrary locations (include $2^r$ combinations). This includes $X = 0$ and $X = U\Psi$ for any $\Psi \in \mathcal{O}_r$ as special examples.

3. Suppose $\Sigma_2 \neq 0$ and $W \neq 0$. Since $\Phi$ and $W$ have orthogonal column spaces, we then require

$$\Theta_2 \Sigma_2^2 \Theta_2 + W^\top W = 0,$$

which further implies

$$\Phi \Sigma_1^2 \Sigma_2 \Theta_2^\top = 0.$$

The solution to the equation above is $\Sigma_2 = 0$, which conflicts with the assumption. This finishes the proof.

The results are identical to those by applying generic theories in Section 2 directly.

## C.2 Proof of Lemma 8

For notational convenience, denote $\widetilde{\mathcal{X}} = \mathcal{X} \backslash \mathcal{U}$. Associate each $X \in \widetilde{\mathcal{X}}$ with a rank deficient set $\mathcal{S} \subseteq [r], \mathcal{S} \neq \emptyset$, which is equivalent with saying that $\Sigma_2 = \Sigma_1 D$, where $D$ is a diagonal matrix with $D_{ii} = 0$ for all $i \in \mathcal{S}$, and $D_{jj} = 1$ for all $j \in \overline{\mathcal{S}} = [r] \backslash \mathcal{S}$. Let $s \in \mathcal{S}$ be the smallest index value in $\mathcal{S}$ and $\overline{s} \in \overline{\mathcal{S}}$ be the smallest index value in $\overline{\mathcal{S}}$.

**Part 1**. We first show that the rank deficient stationary points are strict saddle points, i.e., their eigenvalue satisfies

$$\lambda_{\min}(\nabla^2 \mathcal{F}(X)) \leq -\sigma_s^2(U)$$
$$\lambda_{\max}(\nabla^2 \mathcal{F}(X)) \geq 2\sigma_{\overline{s}}^2(U).$$

If $\overline{\mathcal{S}} = \emptyset$, i.e., $X = 0$, then $\lambda_{\max}(\nabla^2 \mathcal{F}(X)) \geq 0$.

We start with the proof of $\lambda_{\min}(\nabla^2 \mathcal{F}(X))$. Remind that

$$K_X = \begin{bmatrix} X_{(*,1)} X_{(*,1)}^\top & X_{(*,2)} X_{(*,1)}^\top & \cdots & X_{(*,r)} X_{(*,1)}^\top \\ X_{(*,1)} X_{(*,2)}^\top & X_{(*,2)} X_{(*,2)}^\top & \cdots & X_{(*,r)} X_{(*,2)}^\top \\ \vdots & \vdots & \ddots & \vdots \\ X_{(*,1)} X_{(*,r)}^\top & X_{(*,2)} X_{(*,r)}^\top & \cdots & X_{(*,r)} X_{(*,r)}^\top \end{bmatrix}.$$

Let $X_{(*,1)}, \ldots, X_{(*,r)}$ be the columns of $X$. Since $X$ is rank deficient, then there exists a unit vector $w = [w_1, \ldots, w_r]^\top \in \mathbb{R}^r$, $\|w\|_2 = 1$, such that $w^\top X^\top X w = 0$. Let $\phi_s$ be the $s$-th column of $\Phi$, which satisfies $\phi_s^\top X_{(*,i)} = 0$ for any $i \in [r]$ from the construction of $X$, and $z = [z_1^\top, \ldots, z_r^\top]^\top \in \mathbb{R}^{nr}$ be a vector by taking the $i$-th subvector as $z_i = w_{(i)} \phi_s \in \mathbb{R}^n$ for all $i \in [r]$, then

$$\lambda_{\min}(\nabla^2 \mathcal{F}(X)) \leq z^\top \nabla^2 \mathcal{F}(X) z = z^\top (K_X + I_r \otimes XX^\top + X^\top X \otimes I_n - I_r \otimes UU^\top) z$$

$$= \sum_{i,j}^r w_{(i)} w_{(j)} \phi_s^\top X_{(*,j)} X_{(*,i)}^\top \phi_s + \phi_s^\top XX^\top \phi_s + w^\top X^\top X w - \phi_s^\top UU^\top \phi_s$$

$$= 0 + 0 + 0 - \sigma_s^2(U) = -\sigma_s^2(U).$$

The proof of $\lambda_{\max}(\nabla^2 \mathcal{F}(X))$ follows analogous analysis. Let a unit vector $w = [w_1, \ldots, w_r]^\top \in \mathbb{R}^r$ be the singular vector of $X^\top X$ corresponding to the largest singular value $\sigma_{\bar{s}}^2(U)$, and $\phi_{\bar{s}}$ be the $\bar{s}$-th column of $\Phi$, i.e., $\phi_{\bar{s}}^\top XX^\top \phi_{\bar{s}} = \phi_{\bar{s}}^\top UU^\top \phi_{\bar{s}} = \sigma_{\bar{s}}^2(U)$. Let $z = [z_1^\top, \ldots, z_r^\top]^\top \in \mathbb{R}^{nr}$ be a vector by taking the $i$-th subvector as $z_i = w_{(i)} \phi_{\bar{s}} \in \mathbb{R}^n$ for all $i \in [r]$, then

$$\lambda_{\max}(\nabla^2 \mathcal{F}(X)) \geq z^\top \nabla^2 \mathcal{F}(X) z = z^\top (K_X + I_r \otimes XX^\top + X^\top X \otimes I_n - I_r \otimes UU^\top) z$$

$$= \sum_{i,j}^r w_{(i)} w_{(j)} \phi_{\bar{s}}^\top X_{(*,j)} X_{(*,i)}^\top \phi_{\bar{s}} + \phi_{\bar{s}}^\top XX^\top \phi_{\bar{s}} + w^\top X^\top X w - \phi_{\bar{s}}^\top UU^\top \phi_{\bar{s}}$$

$$= \sigma_{\bar{s}}^2(U) + \sigma_{\bar{s}}^2(U) + \sigma_{\bar{s}}^2(U) - \sigma_{\bar{s}}^2(U) = 2\sigma_{\bar{s}}^2(U).$$

When $X = 0$, let $w \in \mathbb{R}^r$ be any unit vector and $\phi \in \mathbb{R}^n$ be a unit vector that satisfies $\phi^\top \Phi = 0$. Construct $z \in \mathbb{R}^{nr}$ as the same way above, then

$$\lambda_{\max}(\nabla^2 \mathcal{F}(X)) \geq z^\top \nabla^2 \mathcal{F}(X) z = z^\top (K_X + I_r \otimes XX^\top + X^\top X \otimes I_n - I_r \otimes UU^\top) z$$

$$= \sum_{i,j}^r w_{(i)} w_{(j)} \phi^\top X_{(*,j)} X_{(*,i)}^\top \phi + \phi^\top XX^\top \phi + w^\top X^\top X w - \phi^\top UU^\top \phi$$

$$= 0 + 0 + 0 - 0 = 0.$$

Next, we show that for any neighborhood $\mathcal{B}$ of $X \in \widetilde{\mathcal{X}}$, there exist $Y_1, Y_2 \in \mathcal{B}$ such that $\mathcal{F}(Y_1) \leq \mathcal{F}(X) \leq \mathcal{F}(Y_2)$. Suppose $X = \Phi \Sigma_1 D \Theta_2$ and $E_1 = \Phi \Sigma_1 D_1 \Theta_2$, where $D + D_1 = I$, then $\langle E_1, X \rangle = 0$. Given $\alpha \in [0, \sqrt{2}]$, let $Y_1 = X + \alpha E_1$, then we have

$$\mathcal{F}(Y_1) = \frac{1}{4} \|Y_1 Y_1^\top - UU^\top\|_F^2 = \frac{1}{4} \left( \|XX^\top - UU^\top\|_F^2 + \alpha^4 \|E_1 E_1^\top\|_F^2 + 2\langle \alpha^2 E_1 E_1^\top, XX^\top - UU^\top \rangle \right)$$

$$= \mathcal{F}(X) + \frac{1}{4} \langle \alpha^2 \Phi \Sigma_1^2 D_1 \Phi^\top, \alpha^2 \Phi \Sigma_1^2 D_1 \Phi^\top - 2\Phi \Sigma_1^2 D_1 \Phi^\top \rangle = \mathcal{F}(X) + \frac{(\alpha^4 - 2\alpha^2)}{4} \Phi \Sigma_1^4 D_1 \Phi^\top$$

$$\leq \mathcal{F}(X).$$

Similarly, let $E_2 = \widetilde{\Phi} \widetilde{\Sigma} \widetilde{\Theta}$, where $\widetilde{\Phi} \in \mathbb{R}^{n \times r}$ has orthogonal columns satisfying $\widetilde{\Phi}^\top \Phi = 0$, $\widetilde{\Sigma} \in \mathbb{R}^{r \times r}$ is any diagonal matrix with nonnegative entries, and $\widetilde{\Theta} \in \mathbb{R}^{r \times r}$ is any orthogonal matrix. Given $\alpha \geq 0$, let $Y_2 = X + \alpha E_2$, then we have

$$\mathcal{F}(Y_2) = \frac{1}{4} \|Y_2 Y_2^\top - UU^\top\|_F^2 = \frac{1}{4} \left( \|XX^\top - UU^\top\|_F^2 + \alpha^4 \|E_2 E_2^\top\|_F^2 \right) \geq \mathcal{F}(X).$$

**Part 2.** Next, we show that for any $X$ in a neighborhood of saddle points, the Hessian matrix $\nabla^2 \mathcal{F}(X)$ has a negative eigenvalue. Given any $X^* \in \widetilde{\mathcal{X}}$ with the associated rank deficient set $\mathcal{S}^* \subseteq [r]$, $\mathcal{S} \neq \emptyset$, let $X = X^* + E$. For any $s \in \mathcal{S}^*$, let $\phi_s$ be the corresponding singular vector of $U$, i.e., the $s$-th column of $\Phi$, $w \in \mathbb{R}^r$ be the singular vector of $X^\top X$ associated with the smallest singular value, and $z \in \mathbb{R}^{nr}$ be a unit vector with the $i$-th subvector as $z_i = w_{(i)} \phi_s$ for all $i \in [r]$, then

$$\begin{aligned}
\lambda_{\min}(\nabla^2 \mathcal{F}(X)) &\leq z^\top \nabla^2 \mathcal{F}(X) z = z^\top (K_X + I_r \otimes XX^\top + X^\top X \otimes I_n - I_r \otimes UU^\top) z \\
&= \sum_{i,j}^r w_{(i)} w_{(j)} \phi_s^\top X_{(*,j)} X_{(*,i)}^\top \phi_s + \phi_s^\top XX^\top \phi_s + w^\top X^\top X w - \phi_s^\top UU^\top \phi_s \\
&= (\phi_s^\top E w^\top)^2 + \phi_s^\top EE^\top \phi_s + \sigma_r^2(X) - \sigma_s^2(U) \\
&\leq 2\|\phi_s^\top E\|_2^2 + \sigma_r^2(X) - \sigma_s^2(U).
\end{aligned} \qquad (22)$$

We claim that from (22), if $\sigma_r(X) \leq \frac{1}{2} \sigma_r(U)$, we have

$$\lambda_{\min}(\nabla^2 \mathcal{F}(X)) \leq -\frac{1}{4} \sigma_r^2(U).$$

The discussion is addressed by the following cases. Let $\mathcal{L}_\Phi$ denote the column space of $\Phi$ and $\mathcal{L}_{\Phi_{\mathcal{S}^*}}$ be the column space of $\Phi_{\mathcal{S}^*}$.

**Case 1:** Suppose $X$ is rank deficient, i.e., $\sigma_r(X) = 0$. Without loss of generality, we can argue that $E$ is also rank deficient. Otherwise, if $E$ is full rank, then there exist some subspace in columns of $X^*$ eliminated by the corresponding subspace in columns of $E$. Therefore, we can consider the rank is deficient in both $X^*$ and $E$ in that particular subspace. Then there exists a subspace $\mathcal{L}_1 \subset \mathcal{L}_\Phi$ such that $E = \mathcal{P}_{\mathcal{L}_1}(E) + (I - \mathcal{P}_{\mathcal{L}_\Phi})(E)$. We can always find a $s \in \mathcal{S}^*$ such that $\phi_s \in \mathcal{L}_\Phi \setminus \mathcal{L}_1$, i.e., $\phi_s^\top \mathcal{P}_{\mathcal{L}_1} x = 0$ for any $x \in \mathbb{R}^n$, such that

$$\phi_s^\top E = \phi_s^\top (\mathcal{P}_{\mathcal{L}_1}(E) + (I - \mathcal{P}_{\mathcal{L}_\Phi})(E)) = 0.$$

This further implies

$$\lambda_{\min}(\nabla^2 \mathcal{F}(X)) \leq -\sigma_s^2(U) \leq -\sigma_r^2(U).$$

**Case 2:** Suppose $X$ has full column rank, and the singular vector $y$ associated with the smallest singular value $\sigma_r(X)$ satisfies $\|\mathcal{P}_{\mathcal{L}_\Phi}(y)\|_2 = 0$ without loss of generality. This implies that for any singular vector $\widetilde{y}$ of $X$, there exists $s \in \mathcal{S}^*$ such that $\phi_s^\top(\widetilde{y}) = 0$. This further implies $\phi_s^\top E = 0$, then combining with (22) we have

$$\lambda_{\min}(\nabla^2 \mathcal{F}(X)) \leq \sigma_r^2(X) - \sigma_s^2(U) \leq -\frac{3}{4} \sigma_r^2(U).$$

**Case 3:** Suppose $X$ has full column rank, and the singular vector $y$ associated with the smallest singular value $\sigma_r(X)$ satisfies $\|\mathcal{P}_{\mathcal{L}_\Phi}(y)\|_2 \in (0, 1]$. This implies that there exists $s \in \mathcal{S}^*$ such that $\|\phi_s^\top E\|_2 \leq \sigma_r(X)$ without loss of generality because there exists a potential subspace of $E$ that is orthogonal to $\phi_s$. If the singular vector associated with smallest singular value of $X$ is not closest to $\phi_s$ for any $s \in \mathcal{S}^* \subset [r]$, then it must be closest to some other $s' \in [r] \setminus \mathcal{S}^*$. Then we can always consider the rank is deficient for $s'$ without loss of generality and the same argument above holds. This further results in

$$\lambda_{\min}(\nabla^2 \mathcal{F}(X)) \leq 3\sigma_r^2(X) - \sigma_s^2(U) \leq -\frac{1}{4} \sigma_r^2(U).$$

## C.3 Proof of Lemma 9

It is obvious that for any $X \in \mathcal{U}$, $\mathcal{F}(X) = 0$, thus it is a global minimum since $\mathcal{F}(Y) \geq 0$ for any $Y \in \mathbb{R}^{n \times r}$. Without loss of generality, let $X = U$, i.e., $\Psi = I$, then we have

$$\nabla^2 \mathcal{F}(U) = K_U + \begin{bmatrix} U_{(*,1)}^\top U_{(*,1)} \cdot I_n & U_{(*,1)}^\top U_{(*,2)} \cdot I_n & \cdots & U_{(*,1)}^\top U_{(*,r)} \cdot I_n \\ U_{(*,2)}^\top U_{(*,1)} \cdot I_n & U_{(*,2)}^\top U_{(*,2)} \cdot I_n & \cdots & U_{(*,2)}^\top U_{(*,r)} \cdot I_n \\ \vdots & \ddots & \vdots & \vdots \\ U_{(*,r)}^\top U_{(*,1)} \cdot I_n & U_{(*,r)}^\top U_{(*,2)} \cdot I_n & \cdots & U_{(*,r)}^\top U_{(*,r)} \cdot I_n \end{bmatrix}.$$

**Part 1**. We first characterize the eigenvectors associated with zero eigenvalues of $\nabla^2 \mathcal{F}(U)$. For any $i$ and $j$ chosen from $1, ..., r$, where $i < j$, we define a vector $v^{(i,j)} \in \mathbb{R}^{nr}$ as

$$v^{(i,j)} = [0^\top, ...., \underbrace{-U_{(*,j)}^\top}_{i\text{-th block}}, ..., \underbrace{U_{(*,i)}^\top}_{j\text{-th block}}, ..., 0^\top]^\top / \sqrt{\|U_{(*,j)}\|_2^2 + \|U_{(*,i)}\|_2^2},$$

where $-U_{(*,j)}$ is the $i$-th block of $v^{(i,j)}$, and $U_{(*,i)}$ is the $j$-th block of $v^{(i,j)}$. Then we can verify

$$v^{(i,j)\top} \nabla^2 \mathcal{F}(U) v^{(i,j)} \cdot \left(\|U_{(*,j)}\|_2^2 + \|U_{(*,i)}\|_2^2\right)$$
$$= U_{(*,j)}^\top U_{(*,j)} \cdot U_{(*,i)}^\top U_{(*,i)} - U_{(*,j)}^\top U_{(*,i)} \cdot U_{(*,i)}^\top U_{(*,j)} - U_{(*,i)}^\top U_{(*,i)} \cdot U_{(*,j)}^\top U_{(*,i)} + U_{(*,i)}^\top U_{(*,i)} \cdot U_{(*,j)}^\top U_{(*,j)}$$
$$+ U_{(*,j)}^\top U_{(*,i)} \cdot U_{(*,i)}^\top U_{(*,j)} - U_{(*,j)}^\top U_{(*,j)} \cdot U_{(*,i)}^\top U_{(*,i)} - U_{(*,i)}^\top U_{(*,i)} \cdot U_{(*,j)}^\top U_{(*,j)} + U_{(*,i)}^\top U_{(*,j)} \cdot U_{(*,j)}^\top U_{(*,i)}$$
$$= 0,$$

which implies that $v^{(i,j)}$ is an eigenvector of $\nabla^2 \mathcal{F}(U)$ and the associated eigenvalue is 0.

We then prove the linear independence among all $v^{(i,j)}$'s by contradiction. Assume that all $v^{(i,j)}$'s are linearly dependent. Then there exist $\alpha_{(i,j)}$'s with at least two nonzero $\alpha_{(i,j)}$'s such that

$$\sum_{i<j} \alpha_{(i,j)} v^{(i,j)} = 0.$$

This further implies that for any $i < k < j$, we have

$$\alpha_{(i,k)} U_{(*,i)} - \alpha_{(k,j)} U_{(*,j)} = 0.$$

Since $U_{(*,j)}$ and $U_{(*,i)}$ are linearly independent, we must have $\alpha_{(i,k)} = \alpha_{(k,j)} = 0$. This is contradicted by our assumption. Thus, all $v^{(i,j)}$'s are linearly independent, i.e., we can obtain all $r(r-1)/2$ eigenvectors associated with zero eigenvalues of $\nabla^2 \mathcal{F}(U)$ by conducting the orthogonalization over all $v^{(i,j)}$. Meanwhile, this also implies that $\mathcal{F}(X)$ is not strongly convex at $X = U$.

We then show that the minimum nonzero eigenvalue of $\nabla^2 \mathcal{F}(U)$ is lower bounded by $\sigma_r^2(U)$. We consider a vector

$$z = [z_1^\top, ..., z_r^\top]^\top \in \mathbb{R}^{nr},$$

which is orthogonal to all $v^{(i,j)}$, i.e., for any $i < j$, we have

$$z_i^\top U_{(*,j)} = z_j^\top U_{(*,i)}.$$

Meanwhile, we also have

$$z^\top \nabla^2 \mathcal{F}(U) z = z^\top (U^\top U \otimes I) z + \sum_{i=1}^r (z_i^\top U_{(*,i)})^2 + 2 \sum_{j<k} (z_j^\top U_{(*,k)})(z_k^\top U_{(*,j)})$$

$$= z^\top (U^\top U \otimes I) z + \sum_{i=1}^r (z_i^\top U_{(*,i)})^2 + 2 \sum_{j<k} (z_j^\top U_{(*,k)})^2$$

$$= z^\top (U^\top U \otimes I) z + z^\top (I \otimes UU^\top) z.$$

We can construct a valid $z$ as follows: let $w = [w_1, ..., w_r]^\top \in \mathbb{R}^r$ be the eigenvector associated with the smallest eigenvalue of $U^\top U$, and y be a vector, which is orthogonal to all $U_{(*,i)}$'s. Then we take

$$z_i = w_{(i)} y.$$

It can be further verified that $z^\top v^{(i,j)} = 0$ for any $(i,j)$, and $z^\top (I \otimes UU^\top) z = 0$. Since both $U^\top U \otimes I$ and $I \otimes UU^\top$ are PSD matrices, then we have from the Weyl's inequality that the minimum nonzero eigenvalue $\lambda_{\min}^+(\nabla^2 \mathcal{F}(U))$ of $\nabla^2 \mathcal{F}(U)$ satisfies

$$\lambda_{\min}^+(\nabla^2 \mathcal{F}(U)) \geq \lambda_{\min}^+(U^\top U \otimes I) = z^\top (U^\top U \otimes I) z = \lambda_{\min}(U^\top U) = \sigma_r^2(U).$$

**Part 2**. Next, we characterize the neighborhood of the global minima. Let $E = X - U$. We then have

$$z^\top (\nabla^2 \mathcal{F}(X) - \nabla^2 \mathcal{F}(U)) z = z^\top (I_r \otimes (UE^\top + EU^\top + EE^\top) + (U^\top E + E^\top U + E^\top E) \otimes I_n + K_E + \widetilde{E}_1 + \widetilde{E}_2) z,$$

where $\widetilde{E}_1$ and $\widetilde{E}_2$ are defined as

$$\widetilde{E}_1 = \begin{bmatrix} E_{(*,1)} U_{(*,1)}^\top & E_{(*,2)} U_{(*,1)}^\top & \cdots & E_{(*,r)} U_{(*,1)}^\top \\ E_{(*,1)} U_{(*,2)}^\top & E_{(*,2)} U_{(*,2)}^\top & \cdots & E_{(*,r)} U_{(*,2)}^\top \\ \vdots & \ddots & \vdots & \vdots \\ E_{(*,1)} U_{(*,r)}^\top & E_{(*,2)} U_{(*,r)}^\top & \cdots & E_{(*,r)} U_{(*,r)}^\top \end{bmatrix}, \quad \widetilde{E}_2 = \begin{bmatrix} U_{(*,1)} E_{(*,1)}^\top & U_{(*,2)} E_{(*,1)}^\top & \cdots & U_{(*,r)} E_{(*,1)}^\top \\ U_{(*,1)} E_{(*,2)}^\top & U_{(*,2)} E_{(*,2)}^\top & \cdots & U_{(*,r)} E_{(*,2)}^\top \\ \vdots & \ddots & \vdots & \vdots \\ U_{(*,1)} E_{(*,r)}^\top & U_{(*,2)} E_{(*,r)}^\top & \cdots & U_{(*,r)} E_{(*,r)}^\top \end{bmatrix}.$$

Meanwhile, we have

$$|z^\top (I_r \otimes (UE^\top + EU^\top + EE^\top)) z| \leq \|UE^\top + EU^\top + EE^\top\|_2 \|z\|_2^2 \leq (2\sigma_1(U)\|E\|_2 + \|E\|_2^2)\|z\|_2^2,$$

$$|z^\top ((U^\top E + E^\top U + E^\top E) \otimes I_n) z| \leq \|U^\top E + E^\top U + E^\top E\|_2 \|z\|_2^2 \leq (2\sigma_1(U)\|E\|_2 + \|E\|_2^2)\|z\|_2^2,$$

$$\left| z^\top (\widetilde{E}_1 + \widetilde{E}_2) z \right| = \left| \sum_{i,j} z_i^\top E_{(*,j)} U_{(*,i)}^\top z_j + \sum_{i,j} z_i^\top U_{(*,j)} E_{(*,i)}^\top z_j \right| = 2 \left| \sum_{i,j} z_i^\top E_{(*,j)} U_{(*,j)}^\top z_i \right| = 2 \left| \sum_i z_i^\top E U^\top z_i \right|$$

$$\leq 2\sigma_1(U) \|E\|_2 \sum_{j=1}^r \|z_i\|_2^2 = 2\sigma_1(U)\|E\|_2 \|z\|_2^2.$$

where the second equality comes from $z_i^\top U_{(*,j)} = z_j^\top U_{(*,i)}$ for all $i,j$'s by constructing $z$ as in Part 1.

We then characterize the eigenvectors associated with negative eigenvalues of $K_E$. For any $i$ and $j$ chosen from $1,...,r$, where $i < j$, we define

$$w^{(i,j)} = [0^\top,...., \underbrace{-E_{(*,j)}^\top}_{i\text{-th block}}, ..., \underbrace{E_{(*,i)}^\top}_{j\text{-th block}} ,..., 0^\top]^\top / \sqrt{\|E_{(*,j)}\|_2^2 + \|E_{(*,i)}\|_2^2},$$

where the $i$-th block of $w^{(i,j)}$ is $-E_{(*,j)}$, and the $j$-th block of $w^{(i,j)}$ is $E_{(*,i)}$. Then we have

$$K_E w^{(i,j)} = \underbrace{\frac{2\left(E_{(*,i)}^\top E_{(*,j)}\right)^2 - 2\|E_{(*,i)}\|_2^2 \|E_{(*,j)}\|_2^2}{\|E_{(*,i)}\|_2^2 + \|E_{(*,j)}\|_2^2}}_{\widetilde{\lambda}} w^{(i,j)},$$

which implies that $w^{(i,j)}$ is an eigenvector of $K_E$ and the associated eigenvalue $\widetilde{\lambda}$ is nonpositive by the Cauchy-Schwarz inequality.

We then prove the linear independence among all $w^{(i,j)}$'s by contradiction. Assume that all $w^{(i,j)}$'s are linearly dependent. Then there exist $\alpha_{(i,j)}$'s with at least two nonzero $\alpha_{(i,j)}$'s such that

$$\sum_{i<j} \alpha_{(i,j)} w^{(i,j)} = 0.$$

This further implies that for any $i < k < j$, we have

$$\alpha_{(i,k)} E_{(*,i)} - \alpha_{(k,j)} E_{(*,j)} = 0.$$

Since $E_{(*,j)}$ and $E_{(*,i)}$ are linearly independent, we must have $\alpha_{(i,k)} = \alpha_{(k,j)} = 0$. This is contradicted by our assumption. Thus, all $w^{(i,j)}$'s are linearly independent, i.e., we can obtain all $r(r-1)/2$ eigenvectors associated with negative eigenvalues of $K_E$ by conducting the orthogonalization over all $w^{(i,j)}$'s.

We consider to construct $z$ analogous to that in Part 1, which is orthogonal to all $w^{(i,j)}$'s. Then we have

$$z^\top w^{(i,j)} = z_i^\top E_{(*,j)} - z_j^\top E_{(*,i)} = 0 \quad \text{for any } i \text{ and } j.$$

This further implies

$$z^\top \begin{bmatrix} E_{(*,1)} E_{(*,1)}^\top & E_{(*,2)} E_{(*,1)}^\top & \cdots & E_{(*,r)} E_{(*,1)}^\top \\ E_{(*,1)} E_{(*,2)}^\top & E_{(*,2)} E_{(*,2)}^\top & \cdots & E_{(*,r)} E_{(*,2)}^\top \\ \vdots & \ddots & \vdots & \vdots \\ E_{(*,1)} E_{(*,r)}^\top & E_{(*,2)} E_{(*,r)}^\top & \cdots & E_{(*,r)} E_{(*,r)}^\top \end{bmatrix} z = \sum_{i,j} z_i^\top E_{(*,j)} E_{(*,i)}^\top z_j = \sum_{i,j} z_i^\top E_{(*,j)} E_{(*,j)}^\top z_i = z^\top (I_r \otimes EE^\top) z.$$

Note that $0 \leq z^\top(I_r \otimes EE^\top)z \leq \sigma_1^2(E)\|z\|_2^2$, which implies $\|K_E\|_2 \leq \sigma_1^2(E)$. Thus, there exists no other eigenvector associated with negative eigenvalues of $K_E$ besides all $w^{(i,j)}$'s. Meanwhile, we also have

$$\lambda_{\min}(K_E) = \min_{i,j} \frac{2(E_{(*,i)}^\top E_{(*,j)})^2 - 2\|E_{(*,i)}\|_2^2\|E_{(*,j)}\|_2^2}{\|E_{(*,i)}\|_2^2 + \|E_{(*,j)}\|_2^2} \geq -\max_{i,j} \frac{2\|E_{(*,i)}\|_2^2\|E_{(*,j)}\|_2^2}{\|E_{(*,i)}\|_2^2 + \|E_{(*,j)}\|_2^2} \geq -\sigma_1^2(E).$$

Combining all results above, we need

$$\|E\|_2 \leq \frac{\sigma_r^2(U)}{8\sigma_1(U)}$$

such that

$$|z^\top(\nabla^2 \mathcal{F}(X) - \nabla^2 \mathcal{F}(U))z| \leq (6\sigma_1(U)\|E\|_2 + 3\|E\|_2^2)\|z\|_2^2 < \frac{4\sigma_r^2(U)}{5}\|z\|_2^2.$$

This implies that

$$z^\top \nabla^2 \mathcal{F}(X) z \geq z^\top \nabla^2 \mathcal{F}(U) z - |z^\top(\nabla^2 \mathcal{F}(X) - \nabla^2 \mathcal{F}(U))z| > \frac{\sigma_r^2(U)}{5}\|z\|_2^2,$$

since $z$ is orthogonal to the eigenvectors corresponding to the zero eigenvalues of $\nabla^2 \mathcal{F}(U)$ by the way of its construction.

## C.4 Proof of Lemma 10

**Part 1**. We first discuss $X \in \mathcal{R}_3'$. Recall that $\nabla \mathcal{F}(X) = (XX^\top - UU^\top)X$. For notational simplicity, let $U = U\Psi_X$, where $\Psi_X = \arg\min_{\Psi \in \mathcal{O}_r} \|X - U\Psi\|_2$.

Let the compact SVD be $X = \Phi_1 \Sigma_1 \Theta_1^\top$, $\Phi_1, \in \mathbb{R}^{n \times r}$, $\Sigma_1, \Sigma_2 \in \mathbb{R}^{r \times r}$. Then we have

$$\|(XX^\top - UU^\top)X\|_F^2 \geq \|(XX^\top - UU^\top)X\|_2^2 \geq \|(XX^\top - UU^\top)\|_2^2 \cdot \sigma_r^2(X). \tag{23}$$

Moreover, we claim that

$$\|XX^\top - UU^\top\|_2^2 \geq 2(\sqrt{2} - 1)\sigma_r^2(U) \cdot \min_{\Psi \in \mathcal{O}_r} \|X - U\Psi\|_2^2. \tag{24}$$

We then demonstrate (24). Let $E = X - U\Psi_X$ with $\Psi_X = \arg\min_{\Psi \in \mathcal{O}_r} \|X - U\Psi\|_2^2$ and the SVD of $U^\top X$ be $U^\top X = A\Sigma B^\top$, then we have $\Psi_X = AB^\top$. This implies

$$X^\top U\Psi_X = B\Sigma B^\top = \Psi_X^\top U^\top X \succeq 0.$$

Further, we have $E^\top U\Psi_X$ is symmetric since

$$E^\top U\Psi_X = X^\top U\Psi_X - \Psi_X^\top U^\top U\Psi_X = \Psi_X^\top U^\top X - \Psi_X^\top U^\top U\Psi_X = \Psi_X^\top U^\top E.$$

Without loss of generality, we assume $\Psi_X = I$, then we have $X^\top U \geq 0$ and $E^\top U = U^\top E$. Substituting $X = U + E$ and denoting $\alpha = 2(\sqrt{2}-1)\sigma_r^2(U)$, we have

$$0 \leq \lambda_{\max}\left(\left(XX^\top - UU^\top\right)^\top \left(XX^\top - UU^\top\right)\right) - \alpha \lambda_{\max}\left((X-U)^\top(X-U)\right)$$

$$\leq \lambda_{\max}\left(\left(XX^\top - UU^\top\right)^\top \left(XX^\top - UU^\top\right) - \alpha(X-U)^\top(X-U)\right)$$

$$= \lambda_{\max}\left(\left(E^\top E\right)^2 + 4E^\top EE^\top U + 2(E^\top U)^2 + 2U^\top UE^\top E - \alpha E^\top E\right)$$

$$= \lambda_{\max}\left(\left(E^\top E + \sqrt{2}E^\top U\right)^2 + (4 - 2\sqrt{2})E^\top EE^\top U + 2U^\top UE^\top E - \alpha E^\top E\right).$$

This implies we only need to show that

$$\lambda_{\max}\left((E^\top E + \sqrt{2}E^\top U)^2 + (4-2\sqrt{2})E^\top EE^\top U + 2U^\top UE^\top E - \alpha E^\top E\right) \geq 0.$$

It is sufficient to show that $(4-2\sqrt{2})E^\top U + 2U^\top U - \alpha I_r \geq 0$. From $E = X - U$ and $X^\top U \geq 0$, we have

$$(4-2\sqrt{2})E^\top U + 2U^\top U - \alpha I_r = (4-2\sqrt{2})X^\top U + 2(\sqrt{2}-1)U^\top U - \alpha I_r \geq 0,$$

provided $2(\sqrt{2}-1)U^\top U - \alpha I_r \geq 0$, which is satisfied by the choice of $\alpha$.

Combining (23), (24), and $\min_{\Psi \in \mathcal{O}_r} \|X - U\Psi\|_2 > \frac{\sigma_r^2(U)}{8\sigma_1(U)}$, we have

$$\|(XX^\top - UU^\top)X\|_F^2 \geq 2(\sqrt{2}-1)\sigma_r^4(U) \cdot \frac{\sigma_r^4(U)}{64\sigma_1^2(U)} \geq \frac{\sigma_r^8(U)}{81\sigma_1^2(U)}.$$

**Part 2.** Next, we discuss $X \in \mathcal{R}_3''$. Let $U = \Phi_1 \Sigma_1 \Theta_1^\top$ and $X = \Phi_2 \Sigma_2 \Theta_2^\top$ be the SVDs, then we have a lower bound of $\|\nabla \mathcal{F}(X)X^\top\|_F$ when $X$ and $U$ has the same column space, i.e,

$$\|\nabla \mathcal{F}(X)X^\top\|_F = \|XX^\top XX^\top - UU^\top XX^\top\|_F = \|\Phi_2 \Sigma_2^4 \Phi_2^\top - \Phi_1 \Sigma_1^2 \Phi_1^\top \Phi_2 \Sigma_2^2 \Phi_2^\top\|_F$$

$$= \sqrt{\|\Sigma_2^4\|_F^2 + \|\Sigma_1^2 \Phi_1^\top \Phi_2 \Sigma_2^2\|_F^2 - 2\text{Tr}(\Phi_1 \Sigma_1^2 \Phi_1^\top \Phi_2 \Sigma_2^6 \Phi_2^\top)}$$

$$\geq \sqrt{\|\Sigma_2^4\|_F^2 + \|\Sigma_1^2 \Sigma_2^2\|_F^2 - 2\text{Tr}(\Sigma_1^2 \Sigma_2^6)} \geq \frac{3}{4}\sqrt{\|\Sigma_2^4\|_F^2} = \frac{3}{4}\|XX^\top XX^\top\|_F, \qquad (25)$$

where the last inequality is from the definition of $\mathcal{R}''$ that $\|\Sigma_2^2\|_F^2 \geq 16\|\Sigma_1^2\|_F^2$ and the minimum is achieved when $(\Sigma_1)_{ii} = \frac{1}{2}(\Sigma_2)_{ii}$ for all $i \in [r]$. Further, we have

$$\|\nabla \mathcal{F}(X)X^\top\|_F \leq \sigma_1(X)\|\nabla \mathcal{F}(X)\|_F \text{ and } \|XX^\top XX^\top\|_F \geq \sigma_1^4(X). \qquad (26)$$

Combining (25) and (26), we have the desired result.

# D Proof of Theorem 5

The proofs are based on the analysis of the general rank $r \geq 1$ case in Theorem 4, combined with the concentration properties of sub-Gaussian matrices $\{A_i\}_{i=1}^d$.

First, we identify the stationary points of $F(X)$ in the following lemma. The proof is provided in Appendix D.1.

**Lemma 11.** For any $X \in \mathcal{U} \cup \{0\}$, $X$ is a stationary point of $F(X)$.

Next, we characterize two types of stationary points. We show in the following lemma that $X = 0$ is the only the strict saddle point, and the Hessian matrix has negative eigenvalues in the neighborhood of $\mathcal{X}$ with high probability if $d$ is large enough. The proof is provided in Appendix D.2

**Lemma 12.** For any $X \in \mathcal{R}_1$, if $\max\{\|XX^\top - UU^\top\|_F^2, \|X\|_F^2, 1\} \leq N_1$ holds for some constant $N_1$ and the number of linear measurements $d$ satisfies $d = \Omega\left(N_1 nr/\sigma_r^2(U)\right)$, then with probability at least $1 - \exp(-C_1 nr)$ for some constant $C_1$, $\nabla^2 F(X)$ contains a negative eigenvalue, i.e.

$$\lambda_{\min}(\nabla^2 F(X)) \leq -\frac{\sigma_r^2(U)}{8}.$$

Moreover, $X = 0$ is a strict saddle point with $\lambda_{\min}(F(0)) \leq -\frac{7}{8}\|U\|_2^2$.

Moreover, we show in the following lemma that any $X \in \mathcal{U}$ is a global minimum, and $F(X)$ is only strongly convex along certain directions in the neighborhood of $X \in \mathcal{U}$ with high probability if $d$ is large enough. The proof is provided in Appendix D.3.

**Lemma 13.** For any $X \in \mathcal{U}$, $X$ is a global minimum, and $\nabla^2 F(X)$ is positive semidefinite. Moreover, for any $X \in \mathcal{R}_2$, if $\max\{\|XX^\top - UU^\top\|_F^2, 4\|U\|_F^2, 1\} \leq N_2$ holds for some constant $N_2$ and $d$ satisfies $d = \Omega\left(N_2 nr/\sigma_r^2(U)\right)$, then with probability at least $1 - \exp(-C_2 nr)$ for some constant $C_2$, we have

$$z^\top \nabla^2 F(X) z \geq \frac{1}{10} \sigma_r^2(U) \|z\|_2^2$$

for any $z \perp \mathcal{E}$, where $\mathcal{E} \subseteq \mathbb{R}^{n \times r}$ is a subspace is spanned by all eigenvectors of $\nabla^2 \mathcal{F}(K_E)$ associated with the negative eigenvalues, where $E = X - U\Psi_X$ and $\Psi_X$ and $K_E$ are defined in (10).

Finally, we show in the following lemma that the gradient $\nabla F(X)$ is sufficiently large norm outside the neighborhood of $\mathcal{X}$ with high probability if $d$ is large enough. The proof is provided in Appendix D.4.

**Lemma 14.** For any $X \in \mathcal{R}_3'$, if $\max\{\|XX^\top - UU^\top\|_F^2, \max_k \|X_{(*,k)}\|_F^2\} \leq N_3$ holds for some constant $N_3$ and $d$ satisfies $d = \Omega\left(N_3 \sqrt{nr} \log(nr) \sigma_1(U)/\sigma_r^4(U)\right)$, then with probability at least $1 - (C_3 nr)^{-1}$ for some constant $C_3$, we have

$$\|\nabla F(X)\|_F > \frac{\sigma_r^4(U)}{18\sigma_1(U)}.$$

Moreover, for any $X \in \mathcal{R}_3''$, if $d = \Omega\left(n\sqrt{r}\log(n)\right)$, then with probability at least $1 - (C_4 n)^{-2}$ for some constant $C_4$, we have

$$\|\nabla F(X)\|_F > \frac{1}{4}\sigma_1^3(X).$$

For $X \in \mathcal{R}_1$, $N_1 \leq (\|XX^\top\|_F + \|UU^\top\|_F)^2 \leq 25\|UU^\top\|_F^2$. Similarly, we have $N_2 \leq 25\|UU^\top\|_F^2$ and $N_3 \leq 25\|UU^\top\|_F^2$. Then combining $\|UU^\top\|_F^2 \leq r\sigma_1^4(U)$ and Lemma 11 – Lemma 14, if $d$ satisfies

$$d = \Omega\left(\max\left\{\frac{\sigma_1^4(U)nr^2}{\sigma_r^2(U)}, \frac{\sigma_1^5(U)r\sqrt{nr}\log(nr)}{\sigma_r^4(U)}, n\sqrt{r}\log(n)\right\}\right),$$

with probability at least $1 - 2\exp(-C_5 nr) - (C_3 nr)^{-1} - (C_4 n)^{-2}$, we have the desired results.

### D.1 Proof of Lemma 11

Recall that the gradient $F(X)$ is

$$\nabla F(X) = \frac{1}{2d} \sum_{i=1}^{d} \langle A_i, XX^\top - UU^\top \rangle \cdot (A_i + A_i^\top)X.$$

It is easy to see that $X \in \mathcal{U} \cup \{0\}$ is a stationary point of $F(X)$. Note that due to the perturbation of the linear mapping $\mathcal{A}$, $X \in \mathcal{X} \setminus \mathcal{U}$ is not a strict saddle point.

### D.2 Proof of Lemma 12

We only need to verify

$$\left|\lambda_{\min}(\nabla^2 F(X)) - \lambda_{\min}(\nabla^2 \mathcal{F}(X))\right| \leq \|\nabla^2 F(X) - \nabla^2 \mathcal{F}(X)\|_2 \leq \frac{\sigma_r^2(U)}{8},$$

where the first inequality is from Weyl's inequality and the second inequality holds with high probability at least $1 - \exp(-cnr)$ if $d = \Omega(N_1 nr/\sigma_r^2(U))$ by taking $\delta = \sigma_r^2(U)/8$ in Lemma 2. Similarly, we have $\lambda_{\min}(\nabla^2 F(0)) \leq -\frac{7}{8}\|U\|_2^2$ with high probability, which finishes the proof.

### D.3 Proof of Lemma 13

First of all, it is easy to see that for any $X \in \mathcal{U}$, $F(X) = 0$ attains the minimal objective value of $F$, thus $X$ is a global minimum. From (18), we have $\nabla^2 F(U) = \text{vec}((A_i + A_i^\top)U) \cdot \text{vec}((A_i + A_i^\top)U)^\top$, which is positive semidefinite.

The rest of the analysis is analogous to the proof of Lemma 12, where we only need to verify

$$\left|\lambda_{\max}(\nabla^2 F(X)) - \lambda_{\max}(\nabla^2 \mathcal{F}(X))\right| \leq \|\nabla^2 F(X) - \nabla^2 \mathcal{F}(X)\|_2 \leq \frac{\sigma_r^2(U)}{10}.$$

Now we only need to verify the bound of $N_2$. Let $\widetilde{\Psi} = \arg\min_{\Psi \in \mathcal{O}_r} \|X - U\Psi\|_2$ and $\widetilde{U} = U\widetilde{\Psi}$, then $\|\widetilde{U}\|_F = \|U\|_F$ and $\sigma_i(\widetilde{U}) = \sigma_i(U)$ for all $i = 1\ldots,r$. From $\min_{\Psi \in \mathcal{O}_r} \|X - U\Psi\|_2 \leq \frac{\sigma_r^2(U)}{8\sigma_1(U)}$, we have

$$\|X - \widetilde{U}\|_F \leq \sqrt{r}\|X - \widetilde{U}\|_2 \leq \sqrt{r}\sigma_r(U) \leq \sqrt{\sum_{i=1}^{r}\sigma_i^2(U)} = \|U\|_F.$$

This implies

$$\|X\|_F \leq \|X - \widetilde{U}\|_F + \|U\|_F \leq 2\|U\|_F.$$

Following the analysis of Lemma 12, we finish the proof.

## D.4 Proof of Lemma 14

**Part 1.** We first discuss $X \in \mathcal{R}_3'$. By taking $\delta = \frac{\sigma_r^4(U)}{18\sigma_1(U)}$ in the analysis of Lemma 2, we have that if $d = \Omega\left(N_3\sqrt{nr}\log(nr)\sigma_1(U)/\sigma_r^4(U)\right)$, then with probability at least $1 - (c_2 nr)^{-1}$,

$$\|\nabla F(X)\|_F \geq \|\nabla \mathcal{F}(X)\|_F - \|\nabla F(X) - \nabla \mathcal{F}(X)\|_F \geq \frac{\sigma_r^4(U)}{18\sigma_1(U)}.$$

**Part 2.** Next, we discuss $X \in \mathcal{R}_3''$. Remind that from (25) we have

$$\|\nabla \mathcal{F}(X)X^\top\|_F \geq \frac{3}{4}\|XX^\top XX^\top\|_F. \tag{27}$$

Moreover, we have

$$\nabla F(X)X^\top = \frac{1}{d}\sum_{i=1}^d \underbrace{\langle A_i, XX^\top - UU^\top\rangle \cdot (A_i + A_i^\top)XX^\top/2}_{\widehat{\Pi}}.$$

Ignore the index $i$ for $\widehat{\Pi}$ for convenience. Consider the $(j,k)$-th entry of $\widehat{\Pi}$, i.e. $\langle A, XX^\top - UU^\top\rangle \cdot (A_{(j,*)} + A_{(*,j)}^\top)XX_{(*,k)}^\top/2$. Analogous to the analysis in Part 1, since $A$ has i.i.d. zero mean sub-Gaussian entries with variance 1, we have $\langle A, XX^\top - UU^\top\rangle$ and $(A_{(j,*)} + A_{(*,j)}^\top)XX_{(*,k)}^\top$ are also zero mean sub-Gaussian entries with variance bounded by $\|XX^\top - UU^\top\|_F^2$ and $\|XX_{(*,k)}^\top\|_F^2$ respectively.

It is easy to check $\mathbb{E}(\nabla F(X)X^\top) = \nabla \mathcal{F}(X)X^\top$. By Lemma 24, we have $\widehat{\Pi}$ is sub-exponential with variance proxy upper bounded by $N_4 = \max\left\{\|XX^\top - UU^\top\|_F^2, \|XX_{(*,k)}^\top\|_F^2\right\}$. Then by the concentration of sub-exponential random variables,

$$\mathbb{P}\left(|(\nabla F(X)X^\top)_{(j,k)} - (\nabla \mathcal{F}(X)X^\top)_{(j,k)}| > t\right) \leq \exp\left(-\frac{c_1 dt}{N_4}\right).$$

This implies

$$\mathbb{P}\left(\|\nabla F(X)X^\top - \nabla \mathcal{F}(X)X^\top\|_F > t\right) \leq n^2 \exp\left(-\frac{c_3 dt}{N_4 n}\right) = \exp\left(-\frac{c_3 dt}{N_4 n} + 2\log n\right). \tag{28}$$

On the other hand, we have

$$\|XX_{(*,k)}^\top\|_F \leq \|XX^\top\|_F,$$

and

$$\|XX^\top - UU^\top\|_F \leq \|XX^\top\|_F + \|UU^\top\|_F \leq 2\|XX^\top\|_F,$$

which implies

$$N_4 = \max\left\{\|XX^\top - UU^\top\|_F^2, \|XX_{(*,k)}^\top\|_F^2\right\} \leq 4\|XX^\top\|_F^2. \tag{29}$$

Let $X = \Psi_X \Sigma_X \Theta_X$ be the SVD of $X$, then

$$\|XX^\top\|_F^2 = \|\Sigma_X^2\|_F^2 = \sum_{i=1}^r \sigma_i^4(X) \leq \sqrt{r \sum_{i=1}^r \sigma_i^8(X)} = \sqrt{r} \|XX^\top XX^\top\|_F. \tag{30}$$

Combining (28), (29), and (30), then if $t = \frac{1}{2}\|XX^\top XX^\top\|_F$ and $d = \Omega\left(n\sqrt{r}\log(n)\right)$, with probability at least $1 - (c_4 n)^{-2}$, we have

$$\|\nabla F(X) X^\top\|_F \geq \|\nabla \mathcal{F}(X) X^\top\|_F - \|\nabla F(X) X^\top - \nabla \mathcal{F}(X) X^\top\|_F \geq \frac{1}{4}\|XX^\top XX^\top\|_F.$$

Combining with

$$\|\nabla F(X) X^\top\|_F \leq \sigma_1(X) \|\nabla F(X)\|_F \text{ and } \|XX^\top XX^\top\|_F \geq \sigma_1^4(X),$$

we have the desired result.

# E  Proof of Corollary 2

For completeness of the analysis, we provide the intermediate results for Corollary 2 as in the analysis for Theorem 5. Recall that for the noisy scenario, we observe

$$y_{(i)} = \langle A_i, M^* \rangle + z_{(i)},$$

where $\{z_{(i)}\}_{i=1}^d$ are independent zero mean sub-Gaussian random noise with variance $\sigma_z^2$. Denoting $M^* = UU^\top$, we have the corresponding objective, gradient, and Hessian matrix as

$$F(X) = \frac{1}{4d} \sum_{i=1}^d \left( \langle A_i, XX^\top - UU^\top \rangle - z_{(i)} \right)^2, \tag{31}$$

$$\nabla F(X) = \frac{1}{2d} \sum_{i=1}^d \left( \langle A_i, XX^\top - UU^\top \rangle - z_{(i)} \right) \cdot (A_i + A_i^\top) X, \text{ and} \tag{32}$$

$$\nabla^2 F(X) = \frac{1}{2d} \sum_{i=1}^d I_r \otimes \left( \langle A_i, XX^\top - UU^\top \rangle - z_{(i)} \right) \cdot (A_i + A_i^\top) + \text{vec}\left((A_i + A_i^\top)X\right) \cdot \text{vec}\left((A_i + A_i^\top)X\right)^\top. \tag{33}$$

We first show the connection between the noisy model and low-rank matrix factorization in the following lemma.

**Lemma 15.** We have $\mathbb{E}(F(X)) = \mathcal{F}(X) + \frac{\sigma_z^2}{4}$, $\mathbb{E}(\nabla F(X)) = \nabla \mathcal{F}(X)$, and $\mathbb{E}(\nabla^2 F(X)) = \nabla^2 \mathcal{F}(X)$.

We have from Lemma 15 that the objective $F(X)$ for noisy model (31) differs from the unbiased estimator of the objective $\mathcal{F}(X)$ for low-rank matrix factorization (9) only by a quantity depending on $\sigma_z$. Moreover, the gradient (32) and the Hessian matrix (33) of the noisy model are unbiased estimators of the counterparts of the low-rank matrix factorization problem in (11) and (12) respectively. These further allow us to derive the lemmas below directly from the counterparts of the low-rank matrix factorization problem in Theorem 4, using the concentrations of sub-Gaussian quantities $\{A_i\}_{i=1}^d$ and $\{z_{(i)}\}_{i=1}^d$. The proofs of the lemmas below are analogous to those of Lemma 11 – Lemma 14, thus we omit them here.

First, we identify the stationary points of $F(X)$ in the following lemma.

**Lemma 16.** For any $X \in \mathcal{U} \cup \{0\}$, $X$ is a stationary point of $F(X)$.

Next, we show in the following lemma that $X = 0$ is the only the strict saddle point, and the Hessian matrix has negative eigenvalues in the neighborhood of $\mathcal{X}$ with high probability if $d$ is large enough.

**Lemma 17.** For any $X \in \mathcal{R}_1$, if $\max\left\{\|XX^\top - UU^\top\|_F^2 + \sigma_z^2, \|X\|_F^2, 1\right\} \leq N_1$ holds for some constant $N_1$ and the number of linear measurements $d$ satisfies $d = \Omega\left(N_1 nr/\sigma_r^2(U)\right)$, then with probability at least $1 - \exp(-C_1 nr)$ for some constant $C_1$, $\nabla^2 F(X)$ contains a negative eigenvalue, i.e.

$$\lambda_{\min}(\nabla^2 F(X)) \leq -\frac{\sigma_r^2(U)}{8}.$$

Moreover, $X = 0$ is a strict saddle point with $\lambda_{\min}(F(0)) \leq -\frac{7}{8}\|U\|_2^2$.

Moreover, we show in the following lemma that any $X \in \mathcal{U}$ is a global minimum, and $F(X)$ is only strongly convex along certain directions in the neighborhood of $X \in \mathcal{U}$ with high probability if $d$ is large enough.

**Lemma 18.** For any $X \in \mathcal{U}$, $X$ is a global minimum, and $\nabla^2 F(X)$ is positive semidefinite. Moreover, for any $X \in \mathcal{R}_2$, if $\max\left\{\|XX^\top - UU^\top\|_F^2 + \sigma_z^2, \|U\|_F^2, 1\right\} \leq N_2$ holds for some constant $N_2$ and $d$ satisfies $d = \Omega\left(N_2 nr/\sigma_r^2(U)\right)$, then with probability at least $1 - \exp(-C_2 nr)$ for some constant $C_2$, we have

$$z^\top \nabla^2 F(X) z \geq \frac{1}{10}\sigma_r^2(U)\|z\|_2^2$$

for any $z \perp \mathcal{E}$, where $\mathcal{E} \subseteq \mathbb{R}^{n \times r}$ is a subspace is spanned by all eigenvectors of $\nabla^2 \mathcal{F}(K_E)$ associated with the negative eigenvalues, where $E = X - U\Psi_X$.

Finally, we show in the following lemma that the gradient $\nabla F(X)$ is sufficiently large norm outside the neighborhood of $\mathcal{X}$ with high probability if $d$ is large enough.

**Lemma 19.** For any $X \in \mathcal{R}_3'$, if $\max\left\{\|XX^\top - UU^\top\|_F^2 + \sigma_z^2, \max_k \|X_{(*,k)}\|_F^2, \sigma_1(U)/\sigma_r^2(U)\right\} \leq N_3$ holds for some constant $N_3$ and $d$ satisfies $d = \Omega\left(N_3 \sqrt{nr}\log(nr)\sigma_1(U)/\sigma_r^4(U)\right)$, then with probability at least $1 - (C_3 nr)^{-1}$ for some constant $C_3$, we have

$$\|\nabla F(X)\|_F > \frac{\sigma_r^4(U)}{18\sigma_1(U)}.$$

Moreover, for any $X \in \mathcal{R}_3''$, if $d = \Omega\left(n\sqrt{r}\log(n)\right)$, then with probability at least $1 - (C_4 n)^{-2}$ for some constant $C_4$, we have

$$\|\nabla F(X)\|_F > \frac{1}{4}\sigma_1^3(X).$$

In terms of the estimation error, the result follows directly from combining Tu et al. (2015) (Lemma 5.3) and Chen and Wainwright (2015) (Corollary 2) for the sub-Gaussian case. Note that $\widehat{X}$ is the optimal solution here. Note that the statistical rate here is consistent with the result for general noisy setting Negahban and Wainwright (2012).

# F   Proof of Theorem 6

First, (p1) follows directly from Lemma 11. It is also immediate that for any $X \in \mathcal{U} \cup \{0\}$, we have $\nabla F(X) = 0$, which implies $X$ is a stationary point of $F(X)$. Moreover, for any $X \in \mathcal{U}$, we have $F(X) = 0$, which implies $X$ is a global minimum.

Then, we have from Candes and Plan (2011); Foucart and Rauhut (2013) that when $A_i$ has i.i.d. zero mean sub-Gaussian entries with variance 1 and $d \geq cnr$, then with high probability, we have that for any matrices $M_1, M_2$ of rank at most $6r$,

$$\left| \frac{1}{d} \sum_{i=1}^d \langle A_i, M_1 \rangle^2 - \|M_1\|_F^2 \right| \leq \rho_1 \|M_1\|_F^2. \tag{34}$$

Note that $\Psi_X = \operatorname{argmin}_{\Psi \in \mathcal{O}_r} \|X - U\Psi_X\|_F^2 = \operatorname{argmin}_{\Psi \in \mathcal{O}_r} \|X - U\Psi_X\|_2^2 = AB^\top$, where the SVD of $U^\top X = A\Sigma B^\top$.

Then we demonstrate (p2). Here we state an intermediate result to be used later.

**Lemma 20** (Lemma 6 in Ge et al. (2017)). *Given $X, U \in \mathbb{R}^{n \times r}$, and $E = X - U\Psi_X$, where $\Psi_X$ is defined in (10), we have $\|EE^\top\|_F^2 \leq 2\|XX^\top - UU^\top\|_F^2$ and $\|E\|_F^2 \leq \frac{1}{2(\sqrt{2}-1)\sigma_r^2(U)}\|XX^\top - UU^\top\|_F^2$.*

Let $z = [E_{(*,1)}^\top, \ldots, E_{(*,r)}^\top,] \in \mathbb{R}^{nr}$, $E = X - U\Psi_X$, and $\Psi_X$ is defined in (10), then

$$z^\top \nabla^2 F(X) z$$

$$= z^\top \left( \frac{1}{2d} \sum_{i=1}^d I_r \otimes \langle A_i, XX^\top - UU^\top \rangle \cdot (A_i + A_i^\top) + \operatorname{vec}((A_i + A_i^\top)X) \cdot \operatorname{vec}((A_i + A_i^\top)X)^\top \right) z$$

$$= \frac{1}{d} \sum_{i=1}^d \left( \langle A_i, EE^\top \rangle^2 - 3\langle A_i, XX^\top - UU^\top \rangle^2 \right) + \frac{2}{d} \sum_{i=1}^d \langle A_i, XX^\top - UU^\top \rangle \cdot \langle (A_i + A_i^\top)X, E \rangle \tag{35}$$

$$\overset{(i)}{\leq} (1 + \rho_1) \|EE^\top\|_F^2 - 3(1 - \rho_1) \|XX^\top - UU^\top\|_F^2 + 4\|\nabla F(X)\|_* \|E\|_2$$

$$\overset{(ii)}{\leq} -\frac{1}{3}\sigma_r^2(U) \|E\|_2^2 + 4\|\nabla F(X)\|_* \|E\|_2$$

where $(i)$ is from (34) and Fenchel's duality theorem, and $(ii)$ is from Lemma 20 by taking $\rho_1 \leq \frac{1}{10}$ and $\|E\|_2 \leq \|E\|_F$.

On the other hand, we have from (35) and Lemma 20 by taking $\rho_1 \leq \frac{1}{10}$ that

$$z^\top \nabla^2 F(X) z \leq (1+\rho_1)\|EE^\top\|_F^2 - 3(1-\rho_1)\|XX^\top - UU^\top\|_F^2 + 4\|\nabla F(X)\|_F \|E\|_F$$
$$\leq -\frac{1}{3}\sigma_r^2(U)\|E\|_F^2 + 4\|\nabla F(X)\|_F \|E\|_F. \tag{36}$$

For $\|\nabla F(X)\|_F \leq \frac{\sigma_r^3(U)}{96}$ and $\|E\|_F \geq \frac{\sigma_r(U)}{4}$, we have from (36) that

$$z^\top \nabla^2 F(X) z \leq -\frac{1}{6}\sigma_r^2(U)\|E\|_F^2,$$

which implies $\lambda_{\min}(\nabla^2 F(X)) \leq -\frac{1}{6}\sigma_r^2(U)$. Since we have

$$\left\{X \mid \|E\|_2 \geq \frac{\sigma_r(U)}{4}\right\} \subseteq \left\{X \mid \|E\|_F \geq \frac{\sigma_r(U)}{4}\right\},$$

then it follows that $\lambda_{\min}(\nabla^2 F(X)) \leq -\frac{1}{6}\sigma_r^2(U)$ also holds in $\mathcal{R}_1$.

To demonstrate (p3), we have the following intermediate results from Tu et al. (2015).

**Lemma 21** (Lemma 5.7 in Tu et al. (2015)). *Given $X, U \in \mathbb{R}^{n \times r}$, and $E = X - U\Psi_X$, where $\Psi_X$ is defined in (10), with $\|E\|_F \leq \frac{\sigma_r(U)}{4}$, then with high probability, we have*

$$\langle \nabla \mathcal{F}(X), E \rangle - \frac{1}{20}\left(\|XX^\top - UU^\top\|_F^2 + \|EX^\top\|_F^2\right) \geq \frac{\sigma_r^2(U)}{4}\|E\|_F^2 + \frac{1}{5}\|XX^\top - UU^\top\|_F^2.$$

**Lemma 22** (Lemma 5.8 in Tu et al. (2015)). *Given $X, U \in \mathbb{R}^{n \times r}$, $E = X - U\Psi_X$, where $\Psi_X$ is defined in (10), with $\|E\|_F \leq \frac{\|U\|_2}{4}$, and any $V \in \mathbb{R}^{n \times r}$, then with high probability, we have*

$$|\langle \nabla \mathcal{F}(X) - \nabla F(X), V \rangle| \leq \frac{1}{10}\|XX^\top - UU^\top\|_F \|VX^\top\|_F.$$

**Lemma 23** (Lemma 5.9 in Tu et al. (2015)). *Given any $X \in \mathbb{R}^{n \times r}$, with high probability, we have*

$$\|XX^\top - UU^\top\|_F^2 \geq \frac{1}{2\|X\|_2}\|\nabla F(X)\|_F^2.$$

Then we have

$$\langle \nabla \mathcal{F}(X), E \rangle = \langle \nabla F(X), E \rangle + \langle \nabla \mathcal{F}(X) - \nabla F(X), E \rangle \overset{(i)}{\leq} \langle \nabla F(X), E \rangle + \frac{1}{10}\|XX^\top - UU^\top\|_F \|EX^\top\|_F$$
$$\overset{(ii)}{\leq} \langle \nabla F(X), E \rangle + \frac{1}{20}\left(\|XX^\top - UU^\top\|_F^1 + \|EX^\top\|_F^2\right) \tag{37}$$

where $(i)$ is from Lemma 22 and $(ii)$ is from the inequality of arithmetic and geometric means. Then we have

$$\langle \nabla F(X), E \rangle \overset{(i)}{\geq} \frac{\sigma_r^2(U)}{4}\|E\|_F^2 + \frac{1}{5}\|XX^\top - UU^\top\|_F^2 \overset{(ii)}{\geq} \frac{\sigma_r^2(U)}{4}\|E\|_F^2 + \frac{1}{10\|X\|_2}\|\nabla F(X)\|_F^2$$
$$\overset{(iii)}{\geq} \frac{\sigma_r^2(U)}{4}\|E\|_F^2 + \frac{1}{20\|U\|_2}\|\nabla F(X)\|_F^2,$$

where $(i)$ is from Lemma 21 and (37), $(ii)$ is from Lemma 23, and $(iii)$ is from $\|X\|_2 \leq \frac{5}{4}\|U\|_2$ given $\|E\|_2 \leq \frac{1}{4}\|U\|_2$.

# G  Further Intermediate Results

## G.1  Proof of Proposition 1

Consider the following regions:

$$\widetilde{\mathcal{R}}_1 \triangleq \left\{ Y \in \mathbb{R}^{n \times r} \mid \sigma_r(Y) \leq \frac{1}{2}\sigma_r(U) \right\},$$

$$\widetilde{\mathcal{R}}_2 \triangleq \left\{ Y \in \mathbb{R}^{n \times r} \mid \min_{\Psi \in \mathcal{O}_r} \|Y - U\Psi\|_2 \leq \frac{\sigma_r^2(U)}{8\sigma_1(U)} \right\}, \text{ and}$$

$$\widetilde{\mathcal{R}}_3 \triangleq \left\{ Y \in \mathbb{R}^{n \times r} \mid \sigma_r(Y) > \frac{1}{2}\sigma_r(U),\ \min_{\Psi \in \mathcal{O}_r} \|Y - U\Psi\|_2 > \frac{\sigma_r^2(U)}{8\sigma_1(U)} \right\}.$$

Then it is obvious to see that $\widetilde{\mathcal{R}}_1 \cup \widetilde{\mathcal{R}}_2 \cup \widetilde{\mathcal{R}}_3 = \mathbb{R}^{n \times r}$. Moreover, we immediately have $\mathcal{R}_1 = \widetilde{\mathcal{R}}_1 \cap \mathcal{R}_3''^{\perp}$ and $\mathcal{R}_3' = \widetilde{\mathcal{R}}_3 \cap \mathcal{R}_3''^{\perp}$. Since for $X \in \mathcal{R}_2$, we have for any $i \in [r]$,

$$|\sigma_i(X) - \sigma_i(U)| \leq \frac{\sigma_r(U)}{8},$$

$\|XX^\top\|_F \leq 2\|UU^\top\|_F$ always holds, i.e., $\mathcal{R}_2 \subseteq \mathcal{R}_3''^{\perp}$, thus $\mathcal{R}_2 = \widetilde{\mathcal{R}}_2 \cap \mathcal{R}_3''^{\perp}$ also holds. Then we have

$$\mathcal{R}_1 \cup \mathcal{R}_2 \cup \mathcal{R}_3' = \left( \widetilde{\mathcal{R}}_1 \cup \widetilde{\mathcal{R}}_2 \cup \widetilde{\mathcal{R}}_3 \right) \cap \mathcal{R}_3''^{\perp} = \mathcal{R}_3''^{\perp}.$$

## G.2  Proof of Proposition 2

For any $\alpha \in (0,1)$ and $\Psi \in \mathcal{O}_r$, $\Psi \neq I_r$, we have

$$\begin{aligned}
\mathcal{F}(\alpha U + (1-\alpha)U\Psi) &= \frac{1}{4}\|(\alpha U + (1-\alpha)U\Psi)(\alpha U + (1-\alpha)U\Psi)^\top - UU^\top\|_F^2 \\
&= \frac{\alpha^2(1-\alpha)^2}{4}\|U(\Psi + \Psi^\top - 2I_r)U^\top\|_F^2 \\
&> 0 = \alpha\mathcal{F}(U) + (1-\alpha)\mathcal{F}(U\Psi).
\end{aligned}$$

## G.3 Proof of Lemma 1

We first demonstrate the objective function. By the definition of $F(X)$, we have

$$\mathcal{F}(X) = \mathbb{E}(F(X)) = \mathbb{E}\left(\frac{1}{4d}\sum_{i=1}^{d}(y_{(i)} - \langle A_i, XX^\top\rangle)^2\right) = \frac{1}{4d}\sum_{i=1}^{d}\mathbb{E}\left(\langle A_i, UU^\top\rangle - \langle A_i, XX^\top\rangle\right)^2$$

$$= \frac{1}{4d}\sum_{i=1}^{d}\mathbb{E}\langle A_i, UU^\top - XX^\top\rangle^2 = \frac{1}{4d}\sum_{i=1}^{d}\mathbb{E}\left(\text{vec}(A_i)^\top \text{vec}(UU^\top - XX^\top)\right)^2$$

$$= \frac{1}{4d}\sum_{i=1}^{d}\mathbb{E}\left(\text{vec}(UU^\top - XX^\top)^\top \text{vec}(A_i)\text{vec}(A_i)^\top \text{vec}(UU^\top - XX^\top)\right)$$

$$= \frac{1}{4}\text{vec}(UU^\top - XX^\top)^\top \cdot \frac{1}{d}\sum_{i=1}^{d}\mathbb{E}(\text{vec}(A_i)\text{vec}(A_i)^\top) \cdot \text{vec}(UU^\top - XX^\top)$$

$$= \frac{1}{4}\|\text{vec}(UU^\top - XX^\top)\|_2^2 = \frac{1}{4}\|UU^\top - XX^\top\|_F^2,$$

Next, we demonstrate the gradient and the Hessian matrix. From the independence of $A_i$'s, we have

$$\mathbb{E}(\nabla F(X)) = \frac{1}{2}\mathbb{E}\left(\langle A_i, XX^\top - UU^\top\rangle \cdot (A_i + A_i^\top)X\right)$$

$$\mathbb{E}(\nabla^2 F(X)) = \frac{1}{2}\mathbb{E}\left(I_r \otimes \langle A_i, XX^\top - UU^\top\rangle \cdot (A_i + A_i^\top) + \text{vec}\left((A_i + A_i^\top)X\right)\cdot\text{vec}\left((A_i + A_i^\top)X\right)^\top\right).$$

We ignore the index $i$ and denote $A_i$ as $A$ for the convenience of notation. The proof is analyzed by entry-wise agreement.

For the $(j,k)$-th entry of gradient $\nabla F(X)$, we have

$$\mathbb{E}(\nabla F(X)_{(j,k)}) = \frac{1}{2}\mathbb{E}\left(\langle A, XX^\top - UU^\top\rangle \cdot (A + A^\top)_{(j,*)}X_{(*,k)}\right)$$

$$= \frac{1}{2}\mathbb{E}\left(\sum_{s,t} A_{(s,t)}(XX^\top - UU^\top)_{(s,t)} \cdot \sum_{l}(A_{(j,l)} + A_{(l,j)})X_{(l,k)}\right)$$

$$\stackrel{(i)}{=} \frac{1}{2}\mathbb{E}\left(\sum_{l} A_{(j,l)}^2 (XX^\top - UU^\top)_{(j,l)}X_{(l,k)} + A_{(l,j)}^2(XX^\top - UU^\top)_{(l,j)}X_{(l,k)}\right)$$

$$= \frac{1}{2}\left(\sum_{l}\mathbb{E}(A_{(j,l)}^2)(XX^\top - UU^\top)_{(j,l)}X_{(l,k)} + \mathbb{E}(A_{(l,j)}^2)(XX^\top - UU^\top)_{(l,j)}X_{(l,k)}\right)$$

$$\stackrel{(ii)}{=} \frac{1}{2}\left(\sum_{l}(XX^\top - UU^\top)_{(j,l)}X_{(l,k)} + (XX^\top - UU^\top)_{(l,j)}X_{(l,k)}\right)$$

$$= (XX^\top - UU^\top)X_{(j,k)}, \tag{38}$$

where $(i)$ is from the independence and zero mean of entries of $A$, and $(ii)$ is from $\sigma^2 = 1$.

We use double index for the Hessian matrix, i.e., denote $(jk, st)$ as the $((k-1)n+j, (t-1)n+s)$-th entry of $\nabla^2 F(X)$. We discuss by separating the two components of $\nabla^2 F(X)$. For the first component,

$$\begin{aligned}
\mathbb{E}\left(\langle A, XX^\top - UU^\top\rangle \cdot (A + A^\top)_{(j,k)}\right) &= \mathbb{E}\left(\sum_{s,t} A_{(s,t)}(XX^\top - UU^\top)_{(s,t)} \cdot (A_{(j,k)} + A_{(k,j)})\right) \\
&= \mathbb{E}\left(A_{(j,k)}^2 (XX^\top - UU^\top)_{(j,k)} + A_{(k,j)}^2 (XX^\top - UU^\top)_{(k,j)}\right) \\
&= 2(XX^\top - UU^\top)_{(j,k)}.
\end{aligned}$$

Therefore, we have

$$\frac{1}{2}\mathbb{E}\left(I_r \otimes \langle A_i, XX^\top - UU^\top\rangle \cdot (A_i + A_i^\top)\right) = I_r \otimes (XX^\top - UU^\top). \tag{39}$$

For the second component

$$\mathbb{E}\left(\text{vec}((A+A^\top)X) \cdot \text{vec}((A+A^\top)X)_{(jk,st)}^\top\right) = \mathbb{E}\left(\text{vec}((A+A^\top)_{(j,*)}X_{(*,k)}) \cdot \text{vec}((A+A^\top)_{(s,*)}X_{(*,t)})^\top\right)$$

$$= \mathbb{E}\left(\left(\sum_l A_{(j,l)}X_{(l,k)} + \sum_m A_{(m,j)}X_{(m,k)}\right) \cdot \left(\sum_l A_{(s,l)}X_{(l,t)} + \sum_m A_{(m,s)}X_{(m,t)}\right)\right).$$

Remind that $K_X = \begin{bmatrix} K_{11} & K_{21} & \cdots & K_{r1} \\ K_{12} & K_{22} & \cdots & K_{r2} \\ \vdots & \vdots & \ddots & \vdots \\ K_{1r} & K_{2r} & \cdots & K_{rr} \end{bmatrix}$, where $K_{kt} = X_{(*,k)}X_{(*,t)}^\top$. If $j \neq s$, we have

$$\begin{aligned}
&\mathbb{E}\left(\text{vec}((A+A^\top)X) \cdot \text{vec}((A+A^\top)X)_{(jk,st)}^\top\right) \\
&= \mathbb{E}(2A_{(j,s)}^2 X_{(s,k)}X_{(j,t)}) = 2X_{(s,k)}X_{(j,t)} = \left(X^\top X \otimes I_n + K_X\right)_{(jk,st)}.
\end{aligned} \tag{40}$$

If $j = s$, we have

$$\begin{aligned}
&\mathbb{E}\left(\text{vec}((A+A^\top)X) \cdot \text{vec}((A+A^\top)X)_{(jk,jt)}^\top\right) \\
&= \mathbb{E}\left(\sum_l A_{(j,l)}^2 X_{(l,k)}X_{(l,t)} + \sum_m A_{(m,j)}^2 X_{(m,k)}X_{(m,t)} + 2A_{(j,j)}^2 X_{(j,k)}X_{(j,t)}\right) \\
&= 2(X_{(*,k)}^\top X_{(*,t)} + X_{(j,k)}X_{(j,t)}) = \left(X^\top X \otimes I_n + K_X\right)_{(jk,jt)}.
\end{aligned} \tag{41}$$

Combining (39), (40), and (41), we have

$$\mathbb{E}(\nabla^2 F(X)_{(jk,st)}) = \nabla^2 \mathcal{F}(X)_{(jk,st)}.$$

### G.4 Proof of Lemma 2

From Lemma 1, we have that $\mathbb{E}(\nabla F(X)) = \nabla \mathcal{F}(X)$ and $\mathbb{E}(\nabla^2 F(X)) = \nabla^2 \mathcal{F}(X)$. We start with an intermediate result to show that the product of two sub-Gaussian random variables is a sub-exponential random variable.

**Lemma 24.** Suppose $X$ and $Y$ are two zero mean sub-Gaussian random variables with variance proxies $\sigma_1^2$ and $\sigma_2^2$ respectively. Let $\sigma^2 = \max\{\sigma_1^2, \sigma_2^2\}$, then $XY$ is a sub-exponential random variable with variance proxy $\sigma^2$, i.e. there exist some constant $c$ such that for all $t > 0$,

$$\mathbb{P}(|XY - \mathbb{E}(XY)| > t) \leq \exp\left(-ct/\sigma^2\right). \quad (42)$$

*Proof.* By the definition of sub-exponential random variables Vershynin (2010), we have that if $Z$ is a centered sub-exponential random variable, we have

$$\|Z\|_{\psi_1} = \sup_{p \geq 1} \frac{1}{p}(\mathbb{E}|Z|^p)^{1/p} = c_1 \sigma_Z^2,$$

where $\|Z\|_{\psi_1}$ is the sub-exponential norm of $Z$ and $\sigma_Z^2$ is the proxy of the variance of $Z$. Using basic inequalities, we have

$$\|\|XY\|\|_{\psi_1} = \sup_{p \geq 1} \frac{1}{p}(\mathbb{E}(XY)^p)^{\frac{1}{p}} \leq \sup_{p \geq 1} p^{-\frac{1}{2}}(\mathbb{E}X^p)^{\frac{1}{p}} p^{-\frac{1}{2}}(\mathbb{E}Y^p)^{\frac{1}{p}} = c_2 \sigma_1 \sigma_2 \leq c_2 \sigma^2.$$

where $c_2$ is a constant and the last equality holds since $X$ and $Y$ are sub-Gaussian random variables. Thus, $XY$ is a sub-exponential random variable with variance proxy $\sigma^2$. Then for general uncentered sub-exponential $XY$, we have that (42) holds for all $t > 0$ for some constant $c$. □

**Part 1:** The perturbation result of the Hessian matrix is discussed first. To bound $\|\nabla^2 F(X) - \nabla^2 \mathcal{F}(X)\|_2$, we first bound $|z^\top \nabla^2 F(X)z - z^\top \nabla^2 \mathcal{F}(X)z|$ for any unit vector $z \in \mathbb{R}^{nr}$, and apply $\varepsilon$-Net argument. Let $z = [z_1^\top, \ldots, z_r^\top,] \in \mathbb{R}^{nr}$ be a unit vector, where $z_i \in \mathbb{R}^n$ for all $i = 1, \ldots, r$, then

$$z^\top \nabla^2 F(X) z = z^\top \left( \frac{1}{2d} \sum_{i=1}^d I_r \otimes \langle A_i, XX^\top - UU^\top \rangle \cdot (A_i + A_i^\top) + \text{vec}((A_i + A_i^\top)X) \cdot \text{vec}((A_i + A_i^\top)X)^\top \right) z$$

$$= \frac{1}{d}\sum_{i=1}^d \underbrace{\frac{1}{2}\sum_{t=1}^r z_t^\top (A_i + A_i^\top) z_t \cdot \langle A_i, XX^\top - UU^\top \rangle}_{\widehat{I}_i} + \frac{1}{d}\sum_{i=1}^d \underbrace{\frac{1}{2}\left( \sum_{t=1}^r z_t(A_i + A_i^\top)X_{(*,t)} \right)^2}_{\widehat{II}_i}. \quad (43)$$

On the other hand,

$$z^\top \nabla^2 \mathcal{F}(X) z = \underbrace{z^\top \left( I_r \otimes (XX^\top - UU^\top) \right) z}_{I} + \underbrace{z^\top \left( X^\top X \otimes I_n + K_X \right) z}_{II}.$$

From the analysis of Lemma 1, we have $\mathbb{E}(\widehat{I}_i) = I$ and $\mathbb{E}(\widehat{II}_i) = II$.

We ignore the index $i$ of $A_i$ for convenience. To bound $\widehat{I}_i$, we have

$$\underbrace{\frac{1}{2}\sum_{t=1}^r z_t^\top (A + A^\top) z_t \cdot \langle A, XX^\top - UU^\top \rangle}_{\widehat{I}} = \underbrace{\sum_{j=1}^n \sum_{k=1}^n \sum_{t=1}^r z_{t(j)} z_{t(k)} A_{(j,k)}}_{\widehat{III}} \cdot \underbrace{\sum_{j=1}^n \sum_{k=1}^n \left( XX^\top - UU^\top \right)_{(j,k)} A_{(j,k)}}_{\widehat{VI}}.$$

Since $A$ has i.i.d. zero mean sub-Gaussian entires with variance 1, then $\widehat{\text{III}}$ is also a zero mean sub-Gaussian with variance upper bounded by 1 since $\|z\|_2 = 1$, and $\widehat{\text{VI}}$ is also a zero mean sub-Gaussian with variance upper bounded by $\|XX^\top - UU^\top\|_F^2$. By Lemma 24, we have each $\widehat{I}_i$ is sub-exponential with proxy $\sigma_1^2 = \max\{1, \|XX^\top - UU^\top\|_F^2\}$. Then, from the concentration of sum of sub-exponential random variables, there exist some constant $c_1$ such that

$$\mathbb{P}\left(\left|\frac{1}{d}\sum_{i=1}^d \widehat{I}_i - I\right| > t_1\right) \leq \exp\left(-\frac{c_1 d t_1}{\sigma_1^2}\right). \tag{44}$$

On the other hand, $\widehat{\text{II}}_i$ is sub-exponential with variance proxy upper bounded by $\sigma_2^2 = \|X\|_F^2$ since $\sum_{t=1}^r z_t(A_i + A_i^\top)X_{(*,t)}$ is a zero mean sub-Gaussian, then from the concentration of sum of sub-exponential random variables, there exist some constant $c_2$ such that

$$\mathbb{P}\left(\left|\frac{1}{d}\sum_{i=1}^d \widehat{\text{II}}_i - \text{II}\right| > t_2\right) \leq \exp\left(-\frac{c_2 d t_2}{\sigma_2^2}\right). \tag{45}$$

Let $t_1 = t_2 = \delta/4$, then combining (43), (44), and (45), for $N_1 \geq \max\{\sigma_1^2, \sigma_2^2\}$, we have

$$\mathbb{P}\left(\left|z^\top(\nabla^2 F(X) - \nabla^2 \mathcal{F}(X))z\right| > \frac{\delta}{2}\right) \leq \exp\left(-\frac{c_3 d \delta}{\sigma_1^2}\right) + \exp\left(-\frac{c_4 d \delta}{\sigma_2^2}\right) \leq 2\exp\left(-\frac{c_5 d \delta}{N_1}\right), \tag{46}$$

Using the $\varepsilon$-Net, we have

$$\|\nabla^2 F(X) - \nabla^2 \mathcal{F}(X)\|_2 = \sup_{z \in \mathbb{R}^{nr}} \left|z^\top(\nabla^2 F(X) - \nabla^2 \mathcal{F}(X))z\right| \leq (1-2\varepsilon)^{-1} \sup_{z \in \mathcal{N}_\varepsilon} \left|z^\top(\nabla^2 F(X) - \nabla^2 \mathcal{F}(X))z\right|. \tag{47}$$

Combining (46) and (47), if we take $\varepsilon = 1/4$, then the covering number of a unit sphere of $\mathbb{R}^{nr}$ can be bounded as $|\mathcal{N}_\varepsilon| \leq 10^{nr} \leq \exp(3nr)$, we have

$$\mathbb{P}\left(\|\nabla^2 F(X) - \nabla^2 \mathcal{F}(X)\|_2 > \delta\right) \leq \mathbb{P}\left(\sup_{z \in \mathcal{N}_{1/4}} \left|z^\top(\nabla^2 F(X) - \nabla^2 \mathcal{F}(X))z\right| > \delta\right)$$

$$\leq 2|\mathcal{N}_{1/4}|\exp\left(-\frac{c_5 d \delta}{N_1}\right) \leq 2\exp\left(3nr - \frac{c_5 d \delta}{N_1}\right).$$

If $d = \Omega(N_1 nr/\delta)$, then with probability at least $1 - \exp(-c_6 nr)$, we have

$$\|\nabla^2 F(X) - \nabla^2 \mathcal{F}(X)\|_2 \leq \delta.$$

**Part 2:** The perturbation result of the gradient is discussed then. Remind that

$$\nabla F(X) = \frac{1}{d}\sum_{i=1}^d \underbrace{\frac{1}{2}\langle A_i, XX^\top - UU^\top\rangle \cdot (A_i + A_i^\top)X}_{\widehat{I}}.$$

Ignore the index $i$ for $\widehat{I}$ for convenience. Consider the $(j,k)$-th entry of $\widehat{I}$, i.e.,

$$\frac{1}{2}\langle A, XX^\top - UU^\top\rangle \cdot (A_{(j,*)} + A_{(*,j)}^\top)X_{(*,k)}.$$

Analogous to the analysis of Part 1, since $A$ has i.i.d. zero mean sub-Gaussian entries with variance 1, we have $\langle A, XX^\top - UU^\top\rangle$ and $(A_{(j,*)} + A_{(*,j)}^\top)X_{(*,k)}$ are also zero mean sub-Gaussian entries with variance bounded by $\|XX^\top - UU^\top\|_F^2$ and $\|X_{(*,k)}\|_F^2$ respectively.

By Lemma 24, we have that $\widehat{I}$ is sub-exponential with variance proxy upper bounded by

$$N_2 \geq \max\{\|XX^\top - UU^\top\|_F^2, \|X_{(*,k)}\|_F^2\}.$$

Then by the concentration of sub-exponential random variables,

$$\mathbb{P}\left(|\nabla F(X)_{(j,k)} - \nabla \mathcal{F}(X)_{(j,k)}| > t\right) \leq \exp\left(-\frac{c_1 dt}{N_2}\right).$$

This implies

$$\mathbb{P}(\|\nabla F(X) - \nabla \mathcal{F}(X)\|_F > t) \leq nr \exp\left(-\frac{c_1 dt}{N_2\sqrt{nr}}\right) = \exp\left(-\frac{c_1 dt}{N_2\sqrt{nr}} + \log(nr)\right).$$

Let $\delta = t$, then if $d = \Omega\left(N_2\sqrt{nr}\log(nr)/\delta\right)$, with probability at least $1 - (c_2 nr)^{-1}$, we have

$$\|\nabla F(X) - \nabla \mathcal{F}(X)\|_F \leq \delta.$$

Combining Part 1 and Part 2, we have the desired result.

### G.5 Proof of Lemma 15

We first demonstrate the objective function. By the definition of $F(X)$, we have

$$\mathcal{F}(X) = \mathbb{E}(F(X)) = \mathbb{E}\left(\frac{1}{4d}\sum_{i=1}^d (y_{(i)} - \langle A_i, XX^\top\rangle)^2\right) = \frac{1}{4d}\sum_{i=1}^d \mathbb{E}\left(\langle A_i, UU^\top\rangle + z_{(i)} - \langle A_i, XX^\top\rangle\right)^2$$

$$= \frac{1}{4d}\sum_{i=1}^d \mathbb{E}\left(\langle A_i, UU^\top - XX^\top\rangle + z_{(i)}\right)^2 = \frac{1}{4d}\sum_{i=1}^d \mathbb{E}\left(\text{vec}(A_i)^\top \text{vec}(UU^\top - XX^\top) + z_{(i)}\right)^2$$

$$\overset{(i)}{=} \frac{1}{4d}\sum_{i=1}^d \mathbb{E}\left(\text{vec}(UU^\top - XX^\top)^\top \text{vec}(A_i)\text{vec}(A_i)^\top \text{vec}(UU^\top - XX^\top) + z_{(i)}^2\right)$$

$$= \frac{1}{4}\text{vec}(UU^\top - XX^\top)^\top \cdot \frac{1}{d}\sum_{i=1}^d \mathbb{E}(\text{vec}(A_i)\text{vec}(A_i)^\top) \cdot \text{vec}(UU^\top - XX^\top) + \frac{\sigma_z^2}{4}$$

$$= \frac{1}{4}\|\text{vec}(UU^\top - XX^\top)\|_2^2 + \frac{\sigma_z^2}{4} = \frac{1}{4}\|UU^\top - XX^\top\|_F^2 + \frac{\sigma_z^2}{4},$$

where $(i)$ from the fact that $z_{(i)}$ has zero mean and is independent of $A_i$.

Next, we demonstrate the gradient and the Hessian matrix. From the independence of $A_i$'s, we have

$$\mathbb{E}(\nabla F(X)) = \frac{1}{2}\mathbb{E}\left(\left(\langle A_i, XX^\top - UU^\top\rangle - z_{(i)}\right)\cdot(A_i + A_i^\top)X\right)$$

$$\mathbb{E}(\nabla^2 F(X)) = \frac{1}{2}\mathbb{E}\left(I_r \otimes \left(\langle A_i, XX^\top - UU^\top\rangle - z_{(i)}\right)\cdot(A_i + A_i^\top) + \text{vec}\left((A_i + A_i^\top)X\right)\cdot\text{vec}\left((A_i + A_i^\top)X\right)^\top\right).$$

We ignore the index $i$ and denote $A_i$ ($z_{(i)}$) as $A$ ($z$) for the convenience of notation. The proof is analyzed by entry-wise agreement.

For the $(j,k)$-th entry of gradient $\nabla F(X)$, we have

$$\mathbb{E}(\nabla F(X)_{(j,k)}) = \frac{1}{2}\mathbb{E}\left(\left(\langle A, XX^\top - UU^\top\rangle - z\right)\cdot(A + A^\top)_{(j,*)}X_{(*,k)}\right)$$

$$\stackrel{(i)}{=} \frac{1}{2}\mathbb{E}\left(\sum_{s,t} A_{(s,t)}(XX^\top - UU^\top)_{(s,t)} \cdot \sum_l (A_{(j,l)} + A_{(l,j)})X_{(l,k)}\right)$$

$$\stackrel{(ii)}{=} (XX^\top - UU^\top)X_{(j,k)}$$

where $(i)$ is from the zero mean of $z$ and $(ii)$ is from (38) in the proof of Lemma 1.

We use double index again for the Hessian matrix, i.e., denote $(jk, st)$ as the $((k-1)n + j, (t-1)n + s)$-th entry of $\nabla^2 F(X)$. We discuss by separating the two components of $\nabla^2 F(X)$. For the first component,

$$\mathbb{E}\left(\left(\langle A, XX^\top - UU^\top\rangle - z\right)\cdot(A + A^\top)_{(j,k)}\right) = \mathbb{E}\left(\sum_{s,t} A_{(s,t)}(XX^\top - UU^\top)_{(s,t)} \cdot (A_{(j,k)} + A_{(k,j)})\right)$$

$$= \mathbb{E}\left(A_{(j,k)}^2(XX^\top - UU^\top)_{(j,k)} + A_{(k,j)}^2(XX^\top - UU^\top)_{(k,j)}\right)$$

$$= 2(XX^\top - UU^\top)_{(j,k)}.$$

The rest of the analysis is identical to that of Lemma 1.